\documentclass[final]{cvpr}

\usepackage{times}
\usepackage{epsfig}
\usepackage{graphicx}
\usepackage{amsmath}
\usepackage{amssymb}
\usepackage{enumitem}
\usepackage{verbatim}
\usepackage{pifont}
\usepackage{multirow, makecell}

\usepackage{dblfloatfix}

\newcommand{\cmark}{\ding{51}}%
\usepackage[pagebackref=true,breaklinks=true,colorlinks,bookmarks=false]{hyperref}

\usepackage{lipsum}

\makeatletter
\newcommand{\printfnsymbol}[1]{%
  \textsuperscript{\@fnsymbol{#1}}%
}
\makeatother
\def\cvprPaperID{582} 
\def\confYear{CVPR 2021}

\begin{document}

\title{MeGA-CDA: Memory Guided Attention for Category-Aware \\ Unsupervised Domain Adaptive Object Detection}
\author{Vibashan VS$^{1}$\hspace{0.1mm}\thanks{Equal contribution} \hspace{0.5mm}\thanks{Work performed during internship at Mercedes-Benz Research and Development India} \hspace{0.5mm},  Vikram Gupta$^{2}$\hspace{0.1mm}\printfnsymbol{1}\hspace{0.0mm}, Poojan Oza$^{1}$\hspace{0.0mm}\printfnsymbol{1}, Vishwanath A. Sindagi$^{1}$\hspace{0.0mm}\printfnsymbol{1},\and
Vishal M. Patel$^{1}$ \\
$^{1}$ Johns Hopkins University, Baltimore, MD, USA \and 
$^{2}$ Mercedes-Benz Research and Development India \\
{\tt\small \{vvishnu2,poza2,vishwanathsindagi,vpatel36\}@jhu.edu,vikram.gupta@daimler.com}
}


\maketitle

\begin{abstract}
\vskip-3.0mm
Existing approaches for unsupervised domain adaptive object detection perform  feature alignment via adversarial training. While these methods achieve reasonable improvements in performance, they typically perform category-agnostic domain alignment,  thereby resulting in negative transfer of features. To overcome this issue, in this work, we attempt to incorporate category information into the domain adaptation process by proposing \textbf{Me}mory \textbf{G}uided \textbf{A}ttention  for \textbf{C}ategory-\textbf{A}ware \textbf{D}omain \textbf{A}daptation (MeGA-CDA). The proposed method consists of employing category-wise discriminators to ensure category-aware feature alignment for learning domain-invariant discriminative features.  However, since the category information is not available for the target samples, we propose to generate memory-guided category-specific attention maps which are then used to route the features  appropriately to the corresponding category discriminator.  The proposed method is evaluated on several benchmark datasets and is shown to outperform existing approaches.

\vskip-5.0mm
\end{abstract}

\section{Introduction}

Object detectors  \cite{viola2001rapid,felzenszwalb2010object,girshick2014rich,girshick2015fast,liu2016ssd,ren2015faster} are a critical part in the inference pipeline of several applications like autonomous navigation, video surveillance, image analysis \etc. Due to this, object detection has received significant interest from the research community. Recent works like  \cite{girshick2014rich,liu2016ssd,ren2015faster} have achieved exceedingly good performance on several benchmark datasets \cite{everingham2010pascal,deng2009imagenet,geiger2013vision,lin2014microsoft}. However, these approaches suffer from severe degradation of performance when evaluated on images that are sampled from a different distribution as compared to that of  training images. Such scenarios are encountered frequently in the real world. For example, consider the case of self-driving cars where the detectors are typically trained on datasets obtained from one particular city or environmental condition (belonging to   source domain) and are expected to be deployed in different city or environment (belonging to target domain). Due to this, it is crucial  to  develop approaches that enable better generalization of detectors.

 \begin{figure}[t!]
	\begin{center}
		\includegraphics[width=0.70\linewidth]{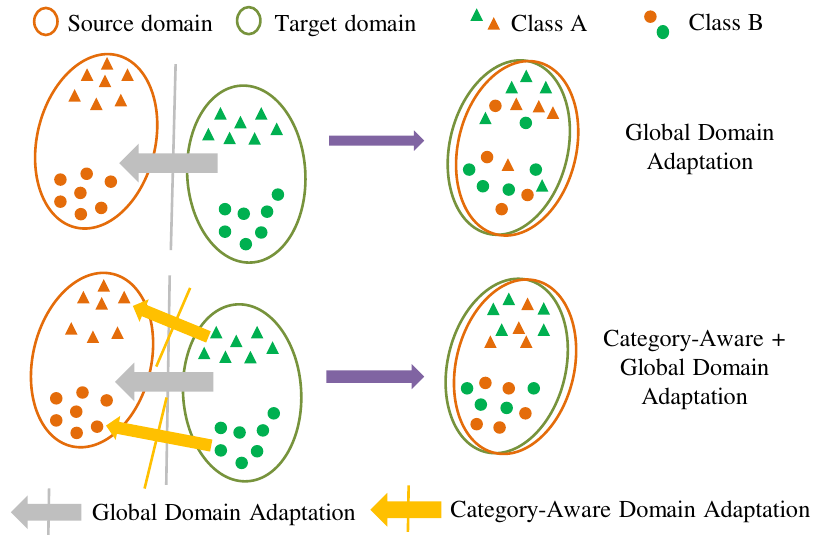}
	\end{center}
 \vskip-7.5pt   \caption{Performing global domain adaptation alone results in potential negative transfer of features. To mitigate this issue, we employ additional category-aware adaptation.}
	\label{fig:comparison} 
	\vskip-6.0mm
\end{figure}

One approach to address this issue is through unsupervised domain adaptation \cite{Chen2018DomainAF,Shan2018,Saito2018StrongWeakDA}, where the goal is to utilize labeled source domain data and unlabeled target domain data to adapt object detector and improve the performance on the unlabeled target domain data. To address this issue, typically these methods attempt to learn domain-invariant features by performing feature alignment between the source and target images. Based on the theoretical insights that minimizing the divergence between the domains reduces the upper-bound error on the target domain \cite{ben2010theory},  they  achieve the feature alignment through adversarial training.  Although these approaches result in considerable improvements, they perform the domain alignment in a category-agnostic way. That is,  they match the global marginal distributions of the two domains without considering the category information. This may lead to cases where the target domain samples are incorrectly aligned with the source-domain samples of a different class (see Fig. \ref{fig:comparison}), thereby resulting in sub-optimal adaptation performance. The task of adapting object detectors is especially prone to this problem due to the presence of multiple categories of objects. 

Considering this issue, we focus on incorporating category information into the domain adaptation process by matching the local joint distribution of features in addition to the global alignment.  In particular, we  perform category-wise  alignment of features by employing category-specific discriminators in the training process. Note that this requires pixel-wise category labels so that the features can be explicitly routed to the respective category-specific discriminator. However, in the case of unsupervised domain adaptation, annotations are not available in any form (bounding boxes/category labels/pixel-wise labels) for the  target dataset. This lack of annotations makes it difficult to use category-specific discriminators.

In order to overcome this challenge, we propose memory-guided attention maps for enabling the category-aware feature alignment. The objective of these attention maps is to focus on objects of specific categories, and hence can be used to route the backbone features  into the appropriate category-specific discriminators. For generating these attention maps, we propose the use of  memory networks \cite{weston2014memory,sukhbaatar2015end,kumar2016ask,kaiser2017learning,fan2019heterogeneous,park2020learning,gong2019memorizing}. During the training process, these memory banks are used to  store prototypes of the objects of different categories, where individual items in the memory correspond to prototypical features of a particular object category. The use of memory network is inspired by their ability to store patterns over longer periods of time. Additionally, the ability to update the patterns using explicit write operations makes them especially useful in domain adversarial training since the features change over the training process.  For determining the attention at a particular location, we use the feature at this location as a query to retrieve relevant items from the different category-specific memory networks. The retrieved items are then compared with the query item and based on the similarity, we compute the category-specific attention map. Furthermore, in order to improve the effectiveness of the memory module and the attention map generation process, we propose a metric-learning based approach that involves learning an appropriate similarity metric based on the available weak-supervision in the source domain. In order to demonstrate the effectiveness of the proposed method, we evaluate it on several benchmark datasets and adaptation protocols. Furthermore, we show that the memory-guided attention maps play an important role in achieving category-wise distribution matching, thereby mitigating the issue of incorrect feature alignment.

To summarize, following are the main contributions of our work:
 \begin{itemize}[topsep=0pt,noitemsep,leftmargin=*]
 	\item We propose memory-guided attention maps for enabling category-wise distribution matching for domain adaptive object detection. 
 	\item In addition, we improve the effectiveness of the memory modules by employing metric learning-based approach for computing the category-specific attention maps. 
 	\item The proposed method is evaluated on several benchmark datasets and is shown to outperform recent domain adaptive detection approaches by a considerable margin. Additionally, we conduct detailed ablation studies to clearly disambiguate the role of memory-guided attention for achieving category-wise alignment. 
 \end{itemize}

\begin{figure*}[t!]
	\begin{center}
		\includegraphics[width=0.70\linewidth]{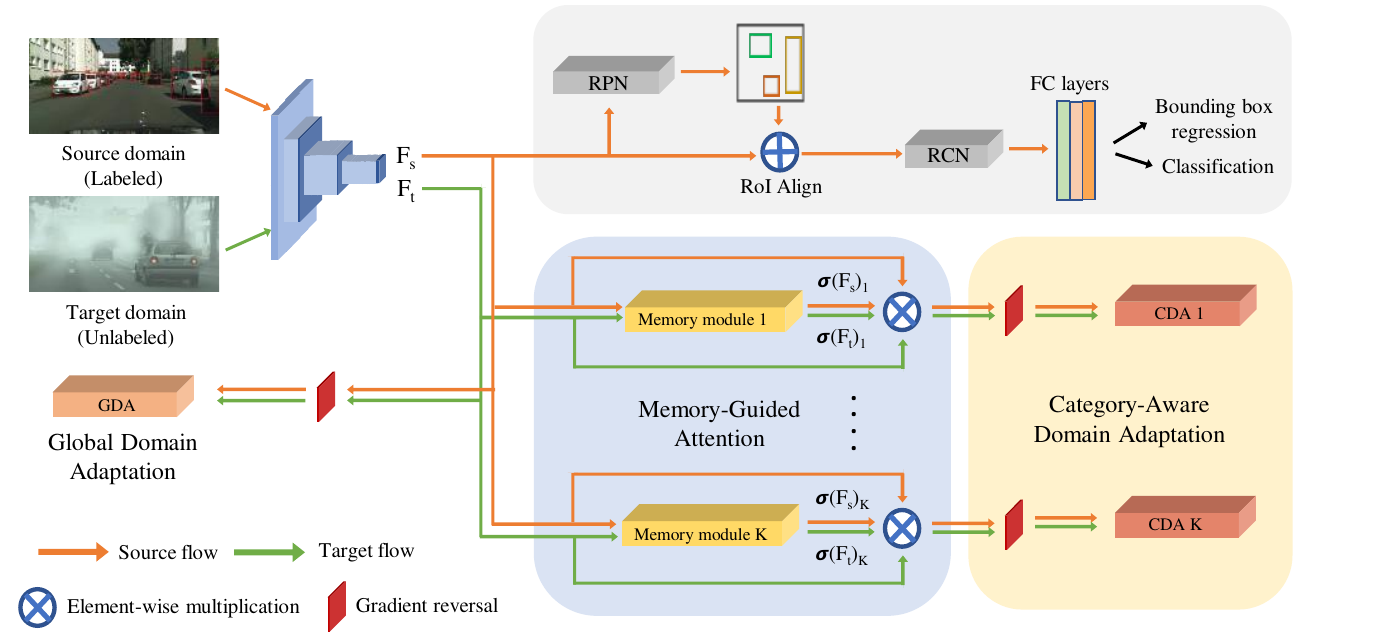}
	\end{center}
 \vskip-7.5pt   \caption{Overview of the proposed approach. Source and target features are aligned through global domain adaptation and category-aware domain adaptation. Global alignment is achieved by category agnostic global discriminator whereas the category-aware alignment is achieved by employing K category-specific discriminators. Since target labels are unavailable, the features to these discriminators are routed  using memory-guided category-specific attention maps. }
	\label{fig:block_diagram} 
	\vskip-9.5pt 
\end{figure*}

\section{Related work}
\label{sec:related_work} 

\noindent{\bf{Object detection}}: The problem of object detection has attracted significant interest due to its widespread applications in several higher-level inference tasks.  
Recent approaches have benefited largely from the success of convolutional neural networks, where different techniques have developed anchor-based strategies for achieving high performance object detection. These approaches can be broadly categorized into (i) two-stage  \cite{ren2015faster} and (ii) 	single-stage approaches  \cite{redmon2016you,liu2016ssd}. 

\noindent \textbf{Unsupervised domain adaptation}: Deep-learning based methods are affected by the domain-shift problem \cite{patel2015visual, wang2018deep}, where networks trained on one distribution of data tend to perform poorly on a different distribution of data. This problem is frequently encountered in the real-world when models are deployed in slightly different conditions compared to the training data. This issue is addressed typically using unsupervised domain adaptation approaches, where the data from different domains are aligned so that the resulting networks/models achieve good generalization performance. Recent domain adaptation approaches involve    feature distribution alignment \cite{tzeng2017adversarial,ganin2014unsupervised,shu2018dirt,saito2018maximum}, residual transfer \cite{long2016unsupervised,long2017deep}, and image-to-image translation approaches \cite{hu2018duplex,murez2018image,hoffman2017cycada,sankaranarayanan2018generate,abavisani2018adversarial,abavisani2016domain,perera2018in2i}. 

\noindent{\bf{Domain adaptation for  object detection}}:  The task of domain adaptation for object detection was recently introduced by Chen \etal \cite{Chen2018DomainAF}, where they address the problem of domain shift at both image-level and instance-level. Shan \etal \cite{Shan2018} proposed to perform joint adaptation at image level using the Cycle-GAN framework \cite{zhu2017unpaired} and at feature level using conventional domain adaptation losses. Saito \etal \cite{Saito2018StrongWeakDA} showed that strong alignment of the features at global level is not necessarily optimal  and   proposed  strong alignment of the local features and weak alignment of the global features.   Kim \etal \cite{kim2019diversify} diversified the labeled data, followed by adversarial learning with the help of multi-domain discriminators. Cai \etal \cite{cai2019exploring} addressed this problem in the semi-supervised setting using mean teacher framework. Zhu \etal \cite{zhu2019adapting} proposed region mining and region-level alignment in order to correctly align the source and target features. Roychowdhury \etal \cite{roychowdhury2019automatic} adapted detectors to a new domain assuming the availability of large number of video data from the target domain. These video data are used to generate pseudo-labels for the target set, which are further employed to train the network. Khodabandeh \etal \cite{khodabandeh2019robust} formulated the domain adaptation training with noisy labels. Specifically, the model is trained on the target domain using a set of noisy bounding boxes that are obtained by a detection model trained only in the source domain. Sindagi \etal \cite{sindagi2020prior} used additional prior about weather into the domain adaptation process.
Hsu \etal \cite{hsu2020every} proposed center-aware feature alignment to emphasize adaptation for foreground regions. Abramov \etal \cite{abramov2020keep} proposed a simple approach by matching the image statistics like color histograms or mean/covariance between the source and target domain. He \etal \cite{he2020domain} proposed an asymmetric tri-way approach to account for the differences in labeling   statistics between source and target domain. Xu \etal \cite{xu2020exploring} added a multi-label classifier as an auxiliary loss to regularize the features. However, the added loss does not pass these category-specific information to the discriminator to help perform the feature alignment. Zhao \etal \cite{zhao2020adaptive} showed that using multi-label classification loss as an auxiliary loss for the domain discriminator yields better performance. Inspired from conditional adversarial networks \cite{long2018conditional}, Zhao \etal \cite{zhao2020adaptive} utilizes the multi-label prediction probability to perform conditional global feature alignment.
\vskip-8.0mm
\section{Proposed method}\label{sec:proposed_method}
\vskip-2.0mm
In this section, we introduce the details of our proposed method. We assume availability of fully-labeled source domain images with bounding-box annotations and unlabeled target domain images without any annotations. For rest of the paper, we denote the source domain dataset as ${D}_s = \{X_s^i, b_s^i, y_s^i\}_{i=1}^{N_s}$, where $X_s^i$ denotes $i^{th}$-image, $b_s^i$ and $y_s^i$ denotes the bounding box annotations and corresponding category labels in the $i^{th}$ source domain image. Also, each category label indicates one of $K$ objects present in the dataset and an extra category for the background classes, i.e., $y_s^i \in \{1,2,..,K+1\}$. Furthermore, we denote the target domain dataset as ${D}_t = \{X_t^i\}_{i=1}^{N_t}$, where $X_t^i$ denotes the $i^{th}$ target domain image. Following  the previous work \cite{Chen2018DomainAF,Shan2018,Saito2018StrongWeakDA,kim2019diversify,khodabandeh2019robust}, we use Faster-RCNN \cite{ren2015faster} as our base model. We denote the backbone feature encoder of the detection network as $\mathcal{E}$. The goal of the  proposed method is to utilize the source domain label information to learn a detection network that can  perform well on the target domain images. To achieve that, we follow a feature alignment approach to match the distribution of features extracted by feature encoder network, for images from source and target dataset through domain adversarial training \cite{ganin2014unsupervised}.

Fig.~\ref{fig:block_diagram} presents an overview of the proposed feature alignment approach which consists of three major modules: \textit{1) Global discriminator} that aligns the entire feature map extracted by the feature encoder network, \textit{2) Category-wise discriminators} that  focus on respective category-specific information to align features belonging to corresponding category between source and target domain. \textit{3) Memory-guided attention mechanism} to enable the training of category-wise discriminators by generating category-specific attention on the extracted feature maps. This attention helps to focus on category information in the extracted feature map for training the respective category-wise discriminators. The attention is generated using a category-specific memory module which stores relevant information for corresponding object category. Details of these modules for feature alignment are  described in  the following sections.

\subsection{Global discriminator for adaptation}
Following the existing works \cite{Chen2018DomainAF,Saito2018StrongWeakDA}, we also employ a global discriminator to perform feature alignment of the feature maps at image-level. The global discriminator, denoted as $\mathcal{D}_{gda}$, takes in the entire feature map extracted from the backbone network and is trained to identify whether the feature map is extracted from source or target domain. More precisely, let us denote a feature map $F_s, \ F_t \in \mathbb{R}^{C \times H \times W}$ extracted from any source and target domain image $X_s$ and $X_t$, respectively. The global discriminator $\mathcal{D}_{gda}$ provides a prediction map of size $H \times W$. The discriminator network is trained with the help of least squared loss supervised with domain label $y_d \in {0, 1}$. For source data, $\forall X_s \in D_s$, and target data, $\forall X_t \in D_t$ the domain labels are set to 1 and 0, respectively. The overall loss function can be written as:
\setlength{\belowdisplayskip}{0pt} \setlength{\belowdisplayshortskip}{0pt}
\setlength{\abovedisplayskip}{0pt} \setlength{\abovedisplayshortskip}{0pt}
\begin{multline}\label{eq:global_align}
\mathcal{L}_{gda}(X_s, X_t) = -\sum_{h=1}^{H} \sum_{w=1}^{W} y_d (1-\mathcal{D}_{gda}(F_s^{(h,w)}))^2 \\
+(1-y_d) (\mathcal{D}_{gda}(F_t^{(h,w)}))^2,
\end{multline}

To match the distribution of the source and the target domain features, we utilize gradient reversal layer as proposed in \cite{ganin2014unsupervised}. The gradient reversal layer flips the gradient sign before propagating the gradients back to the feature extraction network. Hence, the discriminator network $\mathcal{D}_{gda}$ is trained to minimize  Eq.~\ref{eq:global_align} and feature encoder network is trained to maximize Eq.~\ref{eq:global_align}. This adversarial training between feature extractor and discriminator  helps to reduce the domain gap between source and target image features. Furthermore, instead of utilizing binary cross entropy loss for training, we utilize least-squares loss as proposed in \cite{mao2017least} as it   is shown to work better in practice and  helps to stabilize the training process. However, as we argued earlier, the global adaptation is a category-agnostic approach to perform feature alignment between source and target domain. Consequently, this results in negative transfer of features and hurts the overall performance. Hence, using global discriminator alone is not optimal and requires additional strategy to avoid negative transfer of features.

\subsection{Category-wise discriminators for adaptation}
As discussed in the earlier sections, existing methods only consider global feature alignment strategy. In the case of object detection, each image will likely contain multiple categories and hence the feature maps extracted from these images will have features belonging to those respective categories including background features. Hence, addressing negative transfer of features between the categories while aligning source and target domain still remains an important problem in domain adaptive object detection. We address this issue by employing category-wise discriminators (CDA) that focus on aligning respective category-specific features between source and target domains. Specifically, we employ $K$ category-wise discriminators, each focusing on aligning the respective categories. Let us denote discriminator for $k^{th}$ category as $\mathcal{D}^k_{cda}$, $F_s$ and $F_t$ as the features extracted from source and target domain images $X_s$ and $X_t$ respectively. To align the features of the $k^{th}$ category between the source and target domain, we generate attention maps $\sigma(F_s)_k, \ \sigma(F_t)_k \in \{0, 1\}^{H \times W}$ (see Section~\ref{ssec:memory}) to focus on the information related to only $k^{th}$ category. The loss function for category-wise adaptation for $k^{th}$ category can be written as:
\begin{multline}\label{eq:category_align}
\mathcal{L}^k_{cda}(X_s,X_t) = \\ -\sum_{h=1}^{H} \sum_{w=1}^{W} y_d(1-\mathcal{D}^k_{cda}(\sigma(F_s)^{(h,w)}_k \cdot F_s^{(h,w)}))^2 \\
+ (1-y_d) (\mathcal{D}^k_{cda}(\sigma(F_t)^{(h,w)}_k \cdot F_t^{(h,w)}))^2,
\end{multline}
where, $\sigma(\cdot)^{(h,w)}_k=1$ and $\sigma(\cdot)^{(h,w)}_k=0$ indicate the presence and the absence, respectively, of the $k^{th}$ category feature at location $(h,w)$ in the corresponding feature map ($F_s$ or $F_t$), respectively. The major challenge in training with these category-wise discriminator is lack of information regarding the location of category in the feature maps, especially for the target domain data. To this end, we propose a mechanism to predict the attention maps indicating locations of each category with the help of a memory module.

\subsection{Memory-guided attention mechanism}
\label{ssec:memory}
We propose memory-guided attention (MeGA) mechanism to aid the category-wise discriminators in aligning the category-specific features between the source and target domains. Specifically, we employ $K$ memory modules corresponding to the $K$ categories. These memory modules are used to store the class-prototypes of different objects during the training process, so that they can be retrieved for computing the category-specific attention maps. Next, we describe the details regarding memory updates and attention computation.
\vskip-9.5mm


\subsubsection{Memory module}
\vskip-1.5mm
\begin{figure}[t!]
	\begin{center}
		\includegraphics[width=0.80\linewidth]{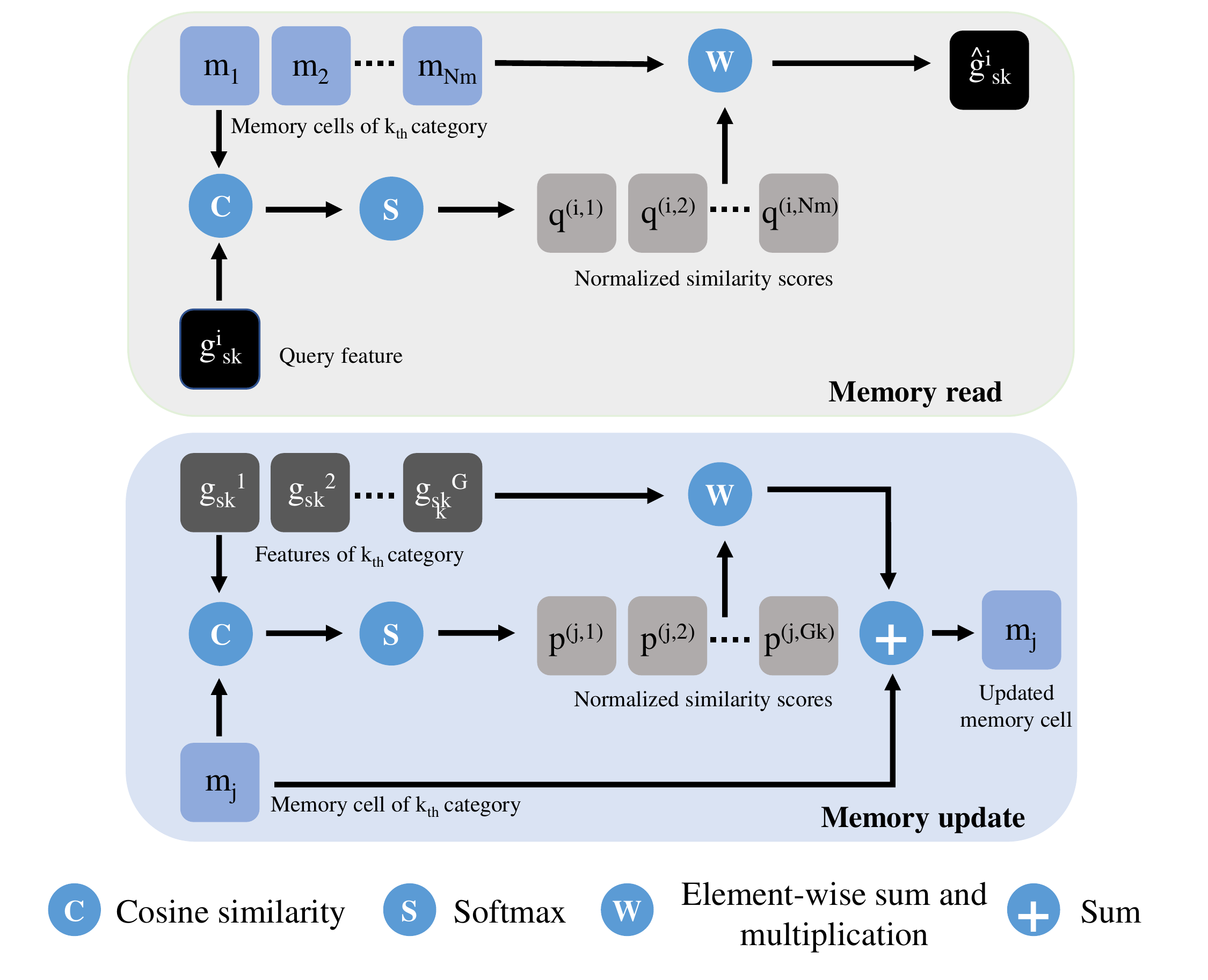}
	\end{center}
	\vskip-10.0pt\caption{Read and write operations for the memory module.}
	\label{fig:memory_module} 
	\vskip-4.5mm
\end{figure}

A memory module has two operations, namely write and read. To write in to the memory, features extracted from the neural network are used to update the memory elements appropriately. Whereas, the memory read operation is used by the features extracted from the neural network to query the memory and retrieve the most similar memory element (or prototypical feature). These operations are illustrated in Fig.~\ref{fig:memory_module}. In the proposed approach, we learn $K$ memory modules, i.e. $M_k \in \mathbb{R}^{N_m \times C}$, corresponding to $K$ categories of the source and target domain. Here, $N_m$ are the number of memory items per category and C are the number of channels in the feature map.\\ 

\vskip-2.5mm
\noindent \textbf{Memory write.} To update the memory elements, we   consider only source domain images as we have access to the bounding box labels to locate the category-specific features in the extracted feature-map $F_s$. For brevity, let us denote $G_k = \{g^i_{s_k} \in \mathbb{R}^{1 \times C}\}_{i=1}^{N_{s_k}}$ all the features belonging to $k^{th}$ category in the feature-map $F_s$. Also, each memory element in memory module $M_k$ is denoted as $m_j \in \mathbb{R}^{1 \times C}$, where $j \in \{1,..., N_m\}$. First, we compute the normalized similarity metric between the memory elements in $M_k$ and the set of features $G_k$ representing $k^{th}$ category as:
\begin{equation}\label{eq:update_memory_1}
p^{(j, i)}=\frac{\exp \left({m}_{j} \cdot {g}_{s_k}^{i}\right)}{\sum_{l \in G_k} \exp \left({m}_{j} \cdot {g}_{s_k}^{l}\right)},
\end{equation}
where, $p$ is an $N_m \times N_k$ similarity map. We utilize this similarity between memory elements and category features to update each memory element using following equation:
\begin{equation}\label{eq:update_memory_2}
{m}^{j} \leftarrow {m}^{j}+\sum_{i \in G_k} p^{(i, j)} {g}_{s_k}^{i}.
\end{equation}
Also, note that if the $k^{th}$ category is not present in the source image, we do not update the elements of the respective memory module $M_k$. Following \cite{park2020learning}, we further regularize the features by making sure that the memory elements should not be too far from the original features. This regularization encourages compactness in the memory module, which reduces intra-class variations. This loss is formulated in the form of L2 distance penalty as:
\setlength{\belowdisplayskip}{1pt} \setlength{\belowdisplayshortskip}{1pt}
\setlength{\abovedisplayskip}{1pt} \setlength{\abovedisplayshortskip}{1pt}
\begin{equation}\label{eq:compact_loss}
\mathcal{L}_{cmp} = \sum_{j=1}^{N_m} \left\|{m}_{j}-{g}^{p}_{s_k}\right\|_{2},
\end{equation}
where, ${g}^{p}_{s_k}$ is a function of $m_j$ and denotes the most similar feature in the set $G_k$ to the memory element $m_j$. In addition to regularizing the memory to be more compact, we enforce a uniqueness constraint to reduce the redundancy in the memory element. Following \cite{park2020learning}, we utilize a triplet loss on the memory elements such that each memory element in the memory module $M_k$ represents unique prototype of the underlying category. This loss can be expressed as follows:
\setlength{\belowdisplayskip}{3pt} \setlength{\belowdisplayshortskip}{3pt}
\setlength{\abovedisplayskip}{3pt} \setlength{\abovedisplayshortskip}{3pt}
\begin{equation}\label{eq:unique_loss}
\mathcal{L}_{unq}= \sum_{j=1}^{N_m} max(\left\|{m}_{j}-{g}^{p}_{s_k}\right\|_{2}-\left\|{m}_{j}-{g}^{n}_{s_k}\right\|_{2}, \alpha),
\end{equation}
where, $\alpha$ denotes the triplet loss margin, ${g}^{p}_{s_k}$ and ${g}^{n}_{s_k}$ denotes respectively the most similar and the second most similar in the feature set $G_k$. Given these constraints, the overall loss for the memory can be defined as:
\setlength{\belowdisplayskip}{3pt} \setlength{\belowdisplayshortskip}{3pt}
\setlength{\abovedisplayskip}{3pt} \setlength{\abovedisplayshortskip}{3pt}
\begin{equation}\label{eq:mem_loss}
\mathcal{L}_{mem}= \mathcal{L}_{cmp} + \mathcal{L}_{unq}.
\end{equation}
%

\noindent \textbf{Memory read.} To retrieve the most similar memory element, we compute the similarity between each item  in the memory $M_k$ and the given query feature. Note that the query feature can be either from the source domain or target domain image, i.e. $g^i_{s_k}$ or $g^i_{t_k}$. For  $g^i_{s_k}$ the normalized similarity is computed using the following equations:
\setlength{\belowdisplayskip}{3pt} \setlength{\belowdisplayshortskip}{3pt}
\setlength{\abovedisplayskip}{3pt} \setlength{\abovedisplayshortskip}{3pt}
\begin{equation}\label{eq:read_memory_1}
q^{(i, j)}=\frac{\exp \left({m}_{j} \cdot {g}_{s_k}^{i}\right)}{\sum_{l \in N_m} \exp \left({m}_{l} \cdot {g}_{s_k}^{i}\right)}.
\end{equation}
Given this normalized similarity $q$, the retrieved feature ${\hat{g}}_{s_k}^{i} \in \mathbb{R}^{1 \times C}$ can be expressed as follows:
\setlength{\belowdisplayskip}{3pt} \setlength{\belowdisplayshortskip}{3pt}
\setlength{\abovedisplayskip}{3pt} \setlength{\abovedisplayshortskip}{3pt}
\begin{equation}\label{eq:read_memory_2}
\hat{{g}}_{s_k}^{i}=\sum_{j \in N_m} q^{(i, j)} {m}_{j}.
\end{equation}
Also, note that we use the same formulation to read from the memory for both the source and target domain features.
\vskip-2.5mm

\begin{table*}[t!]
\caption{Quantitative results (mAP) for Cityscapes $\rightarrow$ Foggy-Cityscapes dataset.}
\vspace{2.5mm}
\huge
\begin{center}
\resizebox{0.6\linewidth}{!}{
\begin{tabular}{|l|cccccccc|c|}
\hline
Method & person & rider & car & truck & bus & train & mcycle & bicycle & mAP\\
\hline\hline
Source Only & 25.8& 33.7& 35.2& 13.0& 28.2& 9.1& 18.7& 31.4& 24.4\\
DAFaster \cite{Chen2018DomainAF} & 25.0& 31.0& 40.5& 22.1& 35.3& 20.2& 20.0& 27.1& 27.6\\
Strong-Weak \cite{Saito2018StrongWeakDA} & 29.9& 42.3& 43.5& 24.5& 36.2& 32.6& 30.0& 35.3& 34.3\\
MAF \cite{he2019multi} & 28.2& 39.5& 43.9& 23.8& 39.9& 33.3& 29.2& 33.9& 34.0\\
D\&Match \cite{kim2019diversify} & 30.8& 40.5& 44.3& 27.2& 38.4& 34.5& 28.4& 32.2& 34.6\\
Selective DA \cite{zhu2019adapting} & 33.5& 38.0& 48.5& 26.5& 39.0& 23.3& 28.0& 33.6& 33.8\\
MTOR \cite{cai2019exploring} & 30.6& 41.4& 44.0& 21.9& 38.6& 40.6& 28.3& 35.6& 35.1\\
ICR-CCR \cite{xu2020exploring} & 32.9& 43.8& 49.2& 27.2& 45.1& 36.4& 30.3& 34.6& 37.4\\
ATF \cite{he2020domain} & 34.6& 47.0& 50.0& 23.7& 43.3& 38.7& 33.4& 38.8& 38.7\\
MCAR \cite{zhao2020adaptive} & 32.0& 42.1& 43.9& \textbf{31.3}& 44.1& 43.4& \textbf{37.4}& 36.6& 38.8\\
 Prior DA \cite{sindagi2020prior} & 36.4& 47.3& 51.7& 22.8& 47.6& 34.1& 36.0& 38.7& 39.3\\
MeGA-CDA (ours) & \textbf{37.7}& \textbf{49.0}& \textbf{52.4}& 25.4& \textbf{49.2}& \textbf{46.9}& 34.5& \textbf{39.0}& \textbf{41.8}\\
\hline 
Oracle \cite{ren2015faster} & 37.2& 48.2& 52.7& 35.2& 52.2& 48.5& 35.3& 38.8& 43.5\\ \hline
\end{tabular}
}
\end{center}
\vspace{-2mm}
\label{foggy_cityscape}
\end{table*}

\subsubsection{Attention mechanism}

\begin{figure}[t!]
	\begin{center}
		\includegraphics[width=1\linewidth]{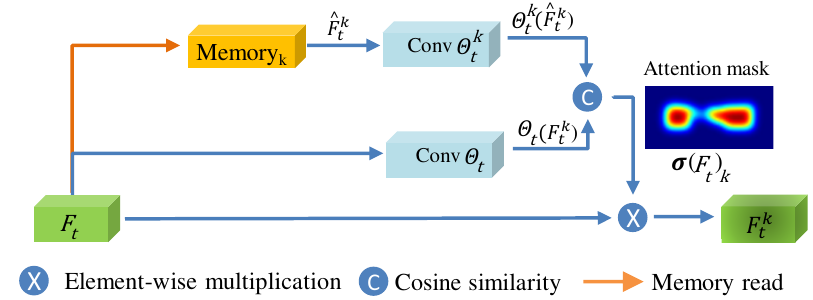}
	\end{center}
	\vskip-7.5pt\caption{The feed-forward path for Memory-guided Attention (MeGA) mechanism. Input source/target feature map is queried to any $k^{th}$-category memory module. Through read operation closest matching elements are retrieved and used to predict attention map through learned similarity. Attention map is used to route the $k$-category information to the $k^{th}$ category discriminator module. }
	\label{fig:attention} 
	\vskip-4.5mm
\end{figure}

We utilize all the memory modules to get attention maps for category-wise discriminators. Specifically, to compute an attention map for the target feature-map $F_t$, we query each element $f_{t} \in \mathbb{R}^{1 \times C}$ to the $k^{th}$ memory module $M_k$ and retrieve a vector $\hat{f}_{t} \in \mathbb{R}^{1 \times C}$ to get a retrieved feature map $\hat{F}^k_t \in \mathbb{R}^{C \times H \times W}$. We compute element-wise similarity between the extracted feature map $F_t$ and the retrieved feature map $\hat{F}^k_t$ to get the attention map for the $k^{th}$ category-wise discriminator $\sigma(F_t)_k$. We explore two choices of similarity function to obtain the attention map.\\
\newline
\noindent \textbf{Cosine similarity.} The most commonly used similarity function in the literature is cosine similarity. We compute the element-wise cosine similarity to get $\sigma(F_t)_k$ of size $H \times W$. It can be expressed as:
\begin{equation}\label{eq:cosine_sim}
\sigma(F_t)_k^{(h, \ w)}=\frac{F^{(h, w)}_t (\hat{F}^{k \ (h,w)}_t)^T}{||F^{(h, w)}_t||_2 ||\hat{F}^{k \ (h, w)}_t||_2},
\end{equation}
\noindent \textbf{Learned similarity.} While the use of cosine similarity to compute the attention maps results in reasonable improvements in accuracy, a closer look at these maps (see Fig. ~\ref{fig:mtl_model} top-row) reveals that the attention generated using cosine similarity is not accurate. To overcome this issue, we explore a similarity metric which is parameterized with a neural network and can be learned during training. In this case, we utilize a metric learning approach where both $F_t$ and $\hat{F}^k_t$ are first passed through a network respectively, $\Theta_t$ and $\Theta^k_t$. To supervise the network we utilize the bounding box information available in the source dataset. In particular, we maximize the cosine similarity between $\Theta_t(F_t)^{(h, w)}$ and $\Theta^k_t(\hat{F}^k_t)^{(h, w)}$ for the location where the category $k$ is present and minimize the similarity where there is an absence of the corresponding category as shown in Fig. \ref{fig:attention}. Then, the attention map can be expressed as:
\begin{equation}\label{eq:cosine_learned}
\sigma(F_t)_k \ = \ Sim(\Theta_t(F_t), \ \Theta^k_t(\hat{F}^k_t)),
\end{equation}
where, $Sim(x, y)$ indicates element-wise cosine similarity between tensor $x$ and $y$ of size $C \times H \times W$, similar to Eq.~\ref{eq:cosine_sim}. The resulting attention $\sigma(\cdot)_k$ is of size $H \times W$. We compute this attention for both source and target images and for all $K$ categories. Before forwarding the attention into the subsequent discriminators, we binarize it with threshold 0.5 \ie, if the normalized similarity is greater than 0.5 we assign 1 to the map and vice versa.

\subsection{Overall training objective for MeGA-CDA}
For our final model training, we add supervised detection loss on the source domain data, which has both images and corresponding bonding box annotations with category labels as described in Sec.~\ref{sec:proposed_method}. We denote the supervised detection loss as $\mathcal{L}_{det}$, which includes both bounding box regression loss and classification loss as described in \cite{ren2015faster}. Including the global, category-wise and memory loss as described in the previous sections, the overall training objective of the proposed method can be expressed as:
\setlength{\belowdisplayskip}{3pt} \setlength{\belowdisplayshortskip}{3pt}
\setlength{\abovedisplayskip}{3pt} \setlength{\abovedisplayshortskip}{3pt}
\begin{multline}\label{eq:overall_objective}
\mathcal{L}^{mega}_{cda} \ = \ \mathcal{L}_{det}(X_s, \ b_s, \ y_s) \ + \ \beta \ \mathcal{L}_{gda}(X_s, \ X_t)\\ + \ \gamma \ \sum_{k=1}^{K} \mathcal{L}^k_{cda}(X_s, \ X_t) \ + \ \lambda \ \mathcal{L}_{mem},
\end{multline}
where, $\beta$, $\gamma$ and $\lambda$ are parameters used to weight the global, category-wise and memory loss, respectively.
\vskip-8.5mm

\section{Experiments and results}
\subsection{Implementation details} We adopt Faster-RCNN~\cite{ren2015faster} network with VGG16 backbone and train using SGD optimizer with learning rate of 0.002 and momentum 0.9 for 6 epochs and then decrease the learning rate to 0.0002. Global and Category-wise discriminators consist of four convolution layers with ReLU non-linearity\footnote{Details of the architecture are included in supplementary material}. The batch size is set to 2 with each batch containing one image from source domain and one from target. We use 20 memory items for each category and each memory item has a dimension of $1\times 1 \times C$, where $C$ denotes the number of channels in the corresponding feature map. The networks $\Theta_t$, $\Theta_t^k$ also consist of 4 convolution layers with ReLU non-linearity. We train the network for 10 epochs and report the mean average precision (mAP) with a threshold of 0.5. The weight of the memory loss, $\lambda$ and of domain adaptors, $\beta$, $\gamma$ are empirically set equal to 0.1 and 0.01, respectively.

\subsection{Quantitative comparison }

In this section, we compare the performance of the proposed method with recent state-of-the-art approaches under three broad categories of adaptation: (i) adverse weather, (ii) synthetic-to-real adaptation, and (iii) cross-camera adaptation. 

\subsubsection{Adverse weather conditions}
Stable object detection performance in different weather conditions is critical for safety critical applications like self-driving cars. Weather conditions introduce image artifacts which can negatively impact the detection performance. To evaluate the effectiveness of proposed method in adverse weather, we utilize Foggy-Cityscapes and Cityscapes as target and source domain respectively.

\noindent \textbf{Dataset:}
The Cityscapes dataset \cite{cordts2016cityscapes} is collected under clear weather conditions and Foggy-Cityscapes \cite{Sakaridis2018SemanticFS} is created by simulating haze on top of the Cityscapes images. Both Cityscapes and Foggy-Cityscapes have 2975 training images and 500 validation images with 8 object categories: \textit{person, rider, car, truck, bus, train, motorcycle and bicycle.} 

\noindent \textbf{Results:}  In Table~\ref{foggy_cityscape}, we report the performance of our framework MeGA-CDA and compare with recent adaptive object detection methods. As it can be observed, MeGA-CDA, outperforms existing approaches by considerable margin,  while improving over the recent best method by an average (absolute) mAP of  2.5\%. Moreover, the proposed method performs consistently well across all categories, demonstrating the  benefits of incorporating category-wise alignment along with global alignment of the features.

\subsubsection{Synthetic data adaptation}
Synthetic data offers an inexpensive alternative to real data collection as it is easier to collect and with appropriate engineering, the synthetic data can be auto-annotated. In spite of the advancements in computer graphics, photo-realistic synthetic data generated using state-of-the-art rendering engines suffer from subtle image artifacts which can result in sub-optimal performance on real-world data.

\noindent \textbf{Dataset:} In this experiment, Sim10k~\cite{johnson2016driving} is used as the source domain and Cityscapes as the target domain. Sim10k has 10,000 images with 58,701 bounding boxes of \textit{car} category, rendered by the gaming engine \textit{Grand Theft Auto}. We use all the Sim10k images for training and evaluate on the bounding boxes of the \textit{car} category from the 500 images of Cityscapes validation set.

\noindent \textbf{Results:}
In Table~\ref{sim10k}, we report the mAP of our framework trained using the Sim10K synthetic data as source and Cityscapes as target. The proposed method, MeGA-CDA,  improves upon the recent best method by 1.8\% mAP (absolute improvement). Since we are adapting from synthetic to real scenario, we observed better alignment when we adapted the features of the third conv layer as well.  Considering that this experiment has only one category of objects, the improvements achieved by the proposed category-aware alignment demonstrates that the memory-guided attention ensures better alignment across the positive and negative (background) class of objects.  

\begin{table}[t!]
\caption{Quantitative results (mAP) for Sim10K $\rightarrow$  Cityscapes.}
\label{sim10k}
\vskip -20pt
\begin{center}
{
\begin{tabular}{|l|c|}
\hline
Method &  mAP \\
\hline\hline
Source Only & 34.3 \\
DAFaster \cite{Chen2018DomainAF} & 38.9 \\
MAF \cite{he2019multi} & 41.1 \\
Strong-Weak \cite{Saito2018StrongWeakDA} & 40.1 \\
ATF \cite{he2020domain} &  42.8\\
Selective DA \cite{zhu2019adapting} &  43.0\\
MeGA-CDA (ours) & \textbf{44.8} \\
\hline
Oracle \cite{ren2015faster} & 62.7 \\ \hline
\end{tabular}
}
\end{center}
\vskip -20pt
\end{table}

\subsubsection{Cross-camera adaptation} Differences in the intrinsic and extrinsic camera properties like resolution, distortion, orientation, location result in images which capture the objects differently from each other in terms of quality, scale and viewing angle. While the collected data can be real, these domain differences will potentially result in severe performance degradation. 

\noindent \textbf{Dataset:} To study this effect of cross-camera domain gap, we conduct two adapataion experiments involving KITTI~\cite{geiger2013vision} and Cityscapes datasets. In the first experiment, we adapt from KITTI to Cityscapes, where as in the second experiment, we adapt from Cityscapes to KITTI. Note that the KITTI dataset consists of 7,481 images,

\noindent \textbf{Results:}
The results of these two experiments are presented in Table~\ref{kitty_city}. In both the experiments, proposed method is able to achieve considerable improvements over the recent best methods. From these results, one may observe that proposed memory-guided category alignment is effective in bridging the domain gap across different camera views and optical properties.


\begin{table}[h!]
\caption{Quantitative results (mAP) for KITTI   $\rightarrow$  Cityscapes  and Cityscapes  $\rightarrow$ KITTI datasets.}
\huge
\begin{center}
\resizebox{.85\linewidth}{!}{
\begin{tabular}{|l|cc|}
\hline
Method & KITTI $\rightarrow$ City & City $\rightarrow$ KITTI \\
\hline\hline
Source Only & 30.2 & 53.5\\
DAFaster \cite{Chen2018DomainAF} & 38.5 & 64.1\\
MAF \cite{he2019multi} & 41.0 & 72.1\\
Strong-Weak \cite{Saito2018StrongWeakDA} & 37.9 & 71.0\\
Selective DA \cite{zhu2019adapting} & 42.5 & -\\
ATF \cite{he2020domain} & 42.1 & 73.5\\
MeGA-CDA (ours) & \textbf{43.0} & \textbf{75.5} \\ \hline
\end{tabular}
}
\end{center}
\label{kitty_city}
\vskip -20pt
\end{table}

\subsection{Ablation studies}
We study the impact of different components of the proposed method, MeGA-CDA, by iteratively adding each module. We use  the  Cityscape$\rightarrow$Foggy-Cityscape adaptation experiment for these ablations. \\

\noindent\textbf{Quantitative analysis}:  The results corresponding to ablation analysis are reported in Table~\ref{ablation_arch_new}. We observe reasonable improvement over the source only baseline by adapting the conv5 features with a global domain discriminator (GDA). By augmenting the global domain discriminators with the proposed memory-guided category-wise discriminators (GDA+CDA+MA) trained with the cosine similarity-based attention, we obtain a further improvement of 0.6\% mAP. This illustrates the benefit of adding category-wise information during domain adaptation. When GDA+CDA+MA is applied at multiple layers (both conv4 and conv5), we observe further improvement of approximately 3\%. Finally, we demonstrate that strengthening the memory module with a metric learned similarity (LS) approach (MeGA-CDA) enhances the capability of the memory banks in capturing the data characteristics and results in further improvements. Specifically, when MeGA-CDA is applied at conv5, we observe an improvement of 2.1\% as compared to GDA+CDA+MA baseline with cosine similarity. Additionally, applying MeGA-CDA at both conv4 and conv5 blocks results in an additional improvement of 2\%. As discussed previously, this sub-network is trained using weak supervision from the ground truth bounding boxes of the source domain and hence, it does not require any additional annotations. \\
\vskip-3.5mm

\begin{table}
\caption{Ablation study on Foggy-Cityscapes. C4 and C5 indicate the adaptation loss at fourth and fifth convolutional block respectively in VGG16 backbone.}
\begin{center}
\huge
\resizebox{1\linewidth}{!}{
\begin{tabular}{|l|cc|cccccccc|c|}
\hline
Method & C5 & C4 & prsn & rider & \multicolumn{1}{c}{car} & \multicolumn{1}{c}{truc} & bus & train & mcycle & bcycle & mAP \\ \hline\hline
Source Only &  &  & 25.8& 33.7& 35.2& 13.0& 28.2& 9.1& 18.7& 31.4& 24.4\\
\hline
GDA & \cmark &  & 35.3 & 44.2 & 51.0 & 23.1 & 44.3 & 28.1 & 27.8 & 37.7 & 36.2 \\ \hline
\multirow{2}{*}{GDA+CDA+MA} & \cmark & & 34.5 & 45.1 & 50.4 & 23.8 & 45.6 & 27.9 & 29.6 & 37.5 & 36.8  \\ 
 & \cmark & \cmark & \textbf{37.8} & 47.1 & \textbf{52.4} & \textbf{29.1} & 48.8 & 29.0 & \textbf{36.7} & \textbf{39.0} & 40.0 \\ \hline
 \multirow{2}{*}{\shortstack[l]{GDA+CDA+MA+LS\\ (MeGA-CDA)}} & \cmark & \textbf{} & 35.9 & 43.7 & 50.8 & 23.4 & 46.5 & \textbf{48.7} & 25.0 & 37.1 & 38.9 \\ 
 & \cmark & \cmark & 37.7 & \textbf{49.0} & \textbf{52.4} & 25.4 & \textbf{49.2} & 46.9 & 34.5 & \textbf{39.0} & \textbf{41.8} \\ \hline
\end{tabular}
}
\end{center}
\label{ablation_arch_new}
\vskip -20pt
\end{table}

\begin{figure}[t!]
\begin{center}
\includegraphics[width=0.47\linewidth]{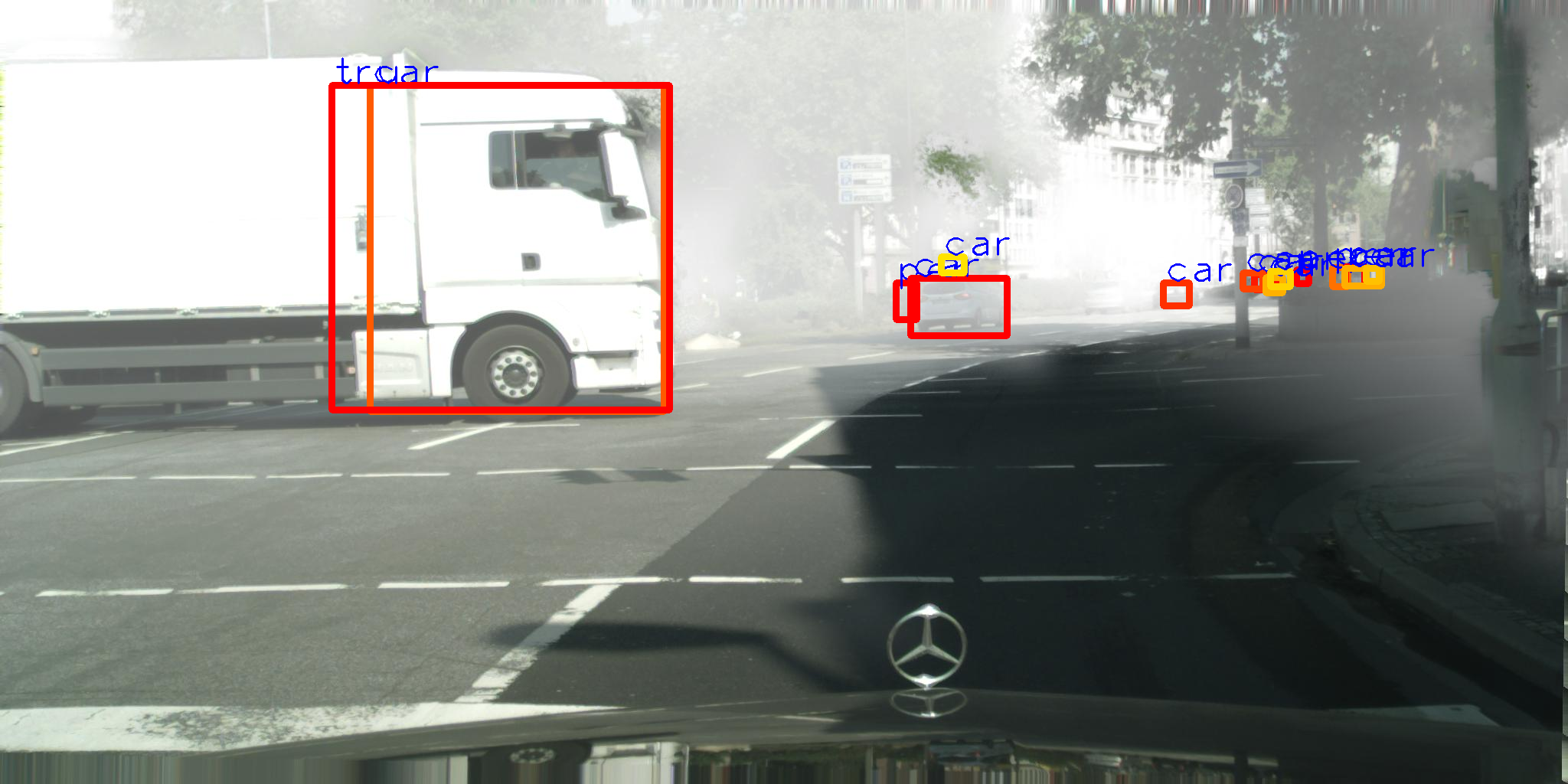}
\includegraphics[width=0.47\linewidth]{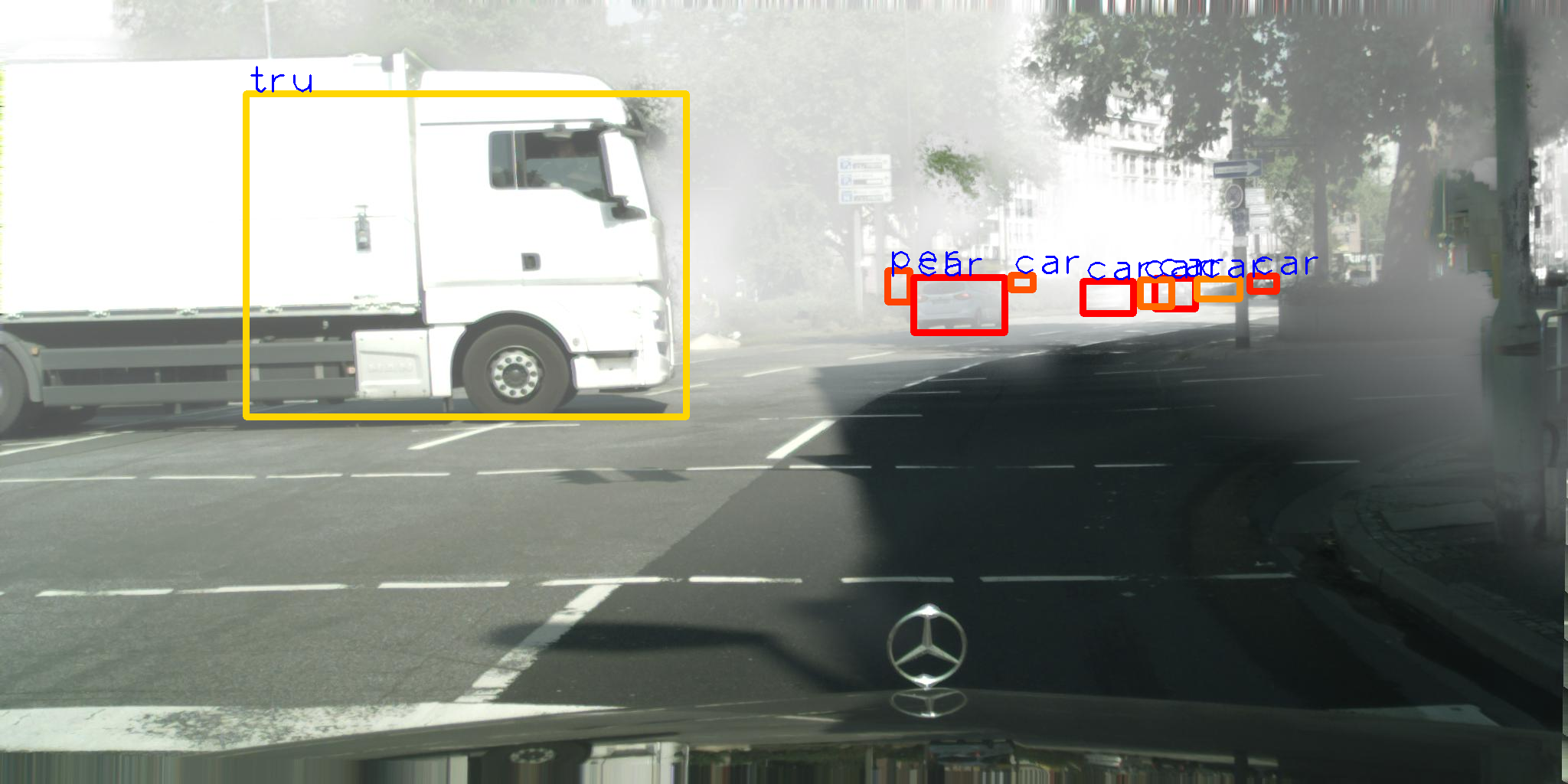}
\includegraphics[width=0.47\linewidth]{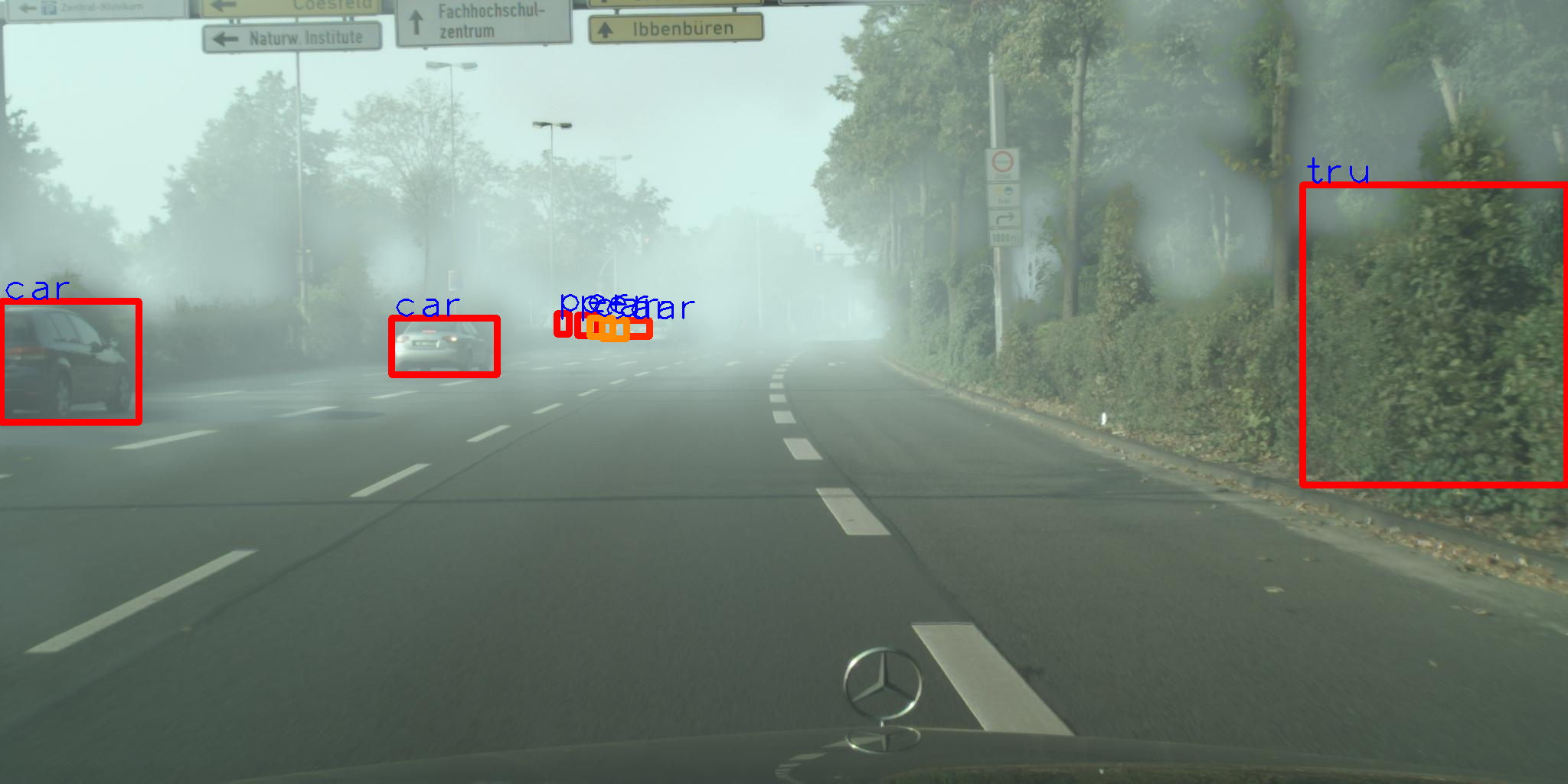}
\includegraphics[width=0.47\linewidth]{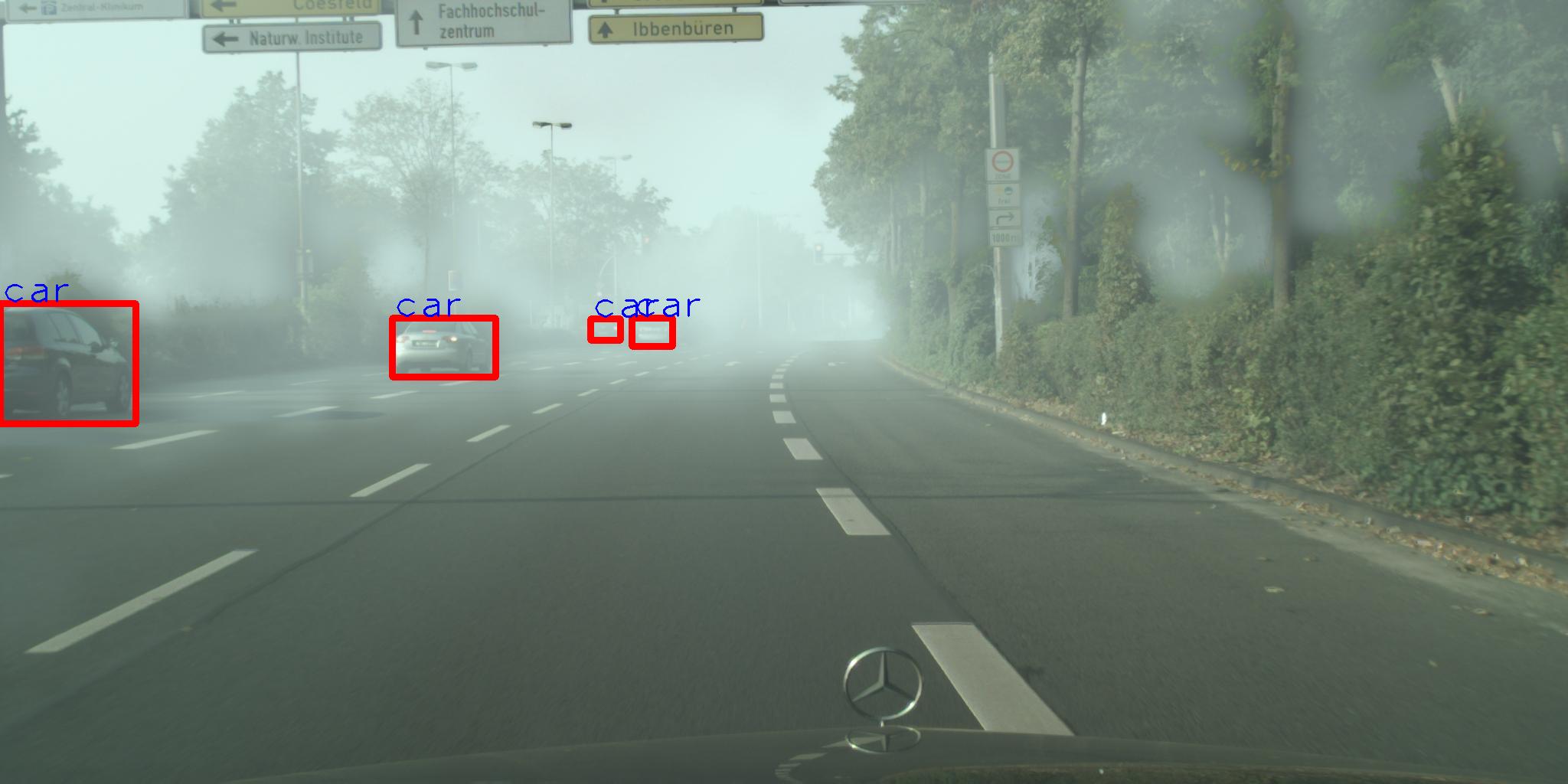}
\includegraphics[width=0.4\linewidth]{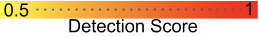}
\vskip -5pt
 GDA \hskip 90pt MeGA-CDA 
\end{center}
\vskip -10.0pt \caption{Qualitative detection results. Global alignment results in miss-detections. In contrast, the proposed approach reduces false-positives while achieving high-quality detections.}
\label{fig:mtl_model_ql}
\vskip-4.5mm
\end{figure}

\begin{figure}[t!]
    \centering
    \includegraphics[width=0.47\textwidth]{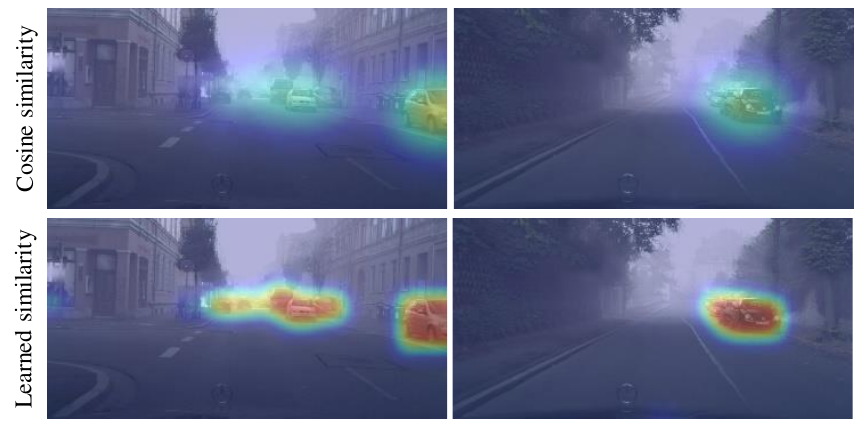}
    \caption{Comparison of attention maps computed using cosine similarity (top-row) and learned similarity based attention (bottom-row). Though cosine similarity based  provides reasonable focus on category features, learned similarity obtains more accurate attention.}
    \label{fig:mtl_model}
\vskip-3.5mm
\end{figure}

\noindent\textbf{Qualitative analysis}: We compare the detections of global alignment approach with proposed category-wise alignment in Fig.~\ref{fig:mtl_model_ql} for the Cityscapes$\rightarrow$Foggy-Cityscapes adaptation experiment. As we can see from  Fig.~\ref{fig:mtl_model_ql}, global alignment-based approach results in errors such as missed-detections (false negatives) or false positives. For example,  background is detected as an object (bottom row) or an object is mistakenly assigned wrong category and bounding box size (top row). The most likely reason for this is negative transfer of features, as global adaptation aligns the feature in category agnostic way. In both cases, the proposed category-wise alignment is able to rectify the error by better countering the negative transfer of features. In Fig.~\ref{fig:mtl_model}, we show attention maps generated for the car category during MeGA-CDA training. For visualization, we overlay the attention maps, generated by the memory module,  on the images. The top row and bottom rows show attention maps computed using cosine similarity and metric learning-based similarity respectively. It can be observed that the cosine similarity-based attention provides reasonable focus on the car category locations. However, with the learned similarity we achieve more effectiveness where the attention spans majority of the car region. This is expected as learned similarity is trained through metric learning with weak supervision from the source domain ground truth and thus results in  guided learning of memory items as well.
\vskip-2.5mm

\section{Conclusions}
\vskip-1.5mm
We presented a category-aware feature alignment approach for domain adaptive object detection. Specifically, we incorporate category information into the domain alignment process by introducing category-aware discriminators. To overcome the issue of lack of category labels, especially in target domain, we propose memory-guided attention mechanism that generate category-specific attention maps for routing the features into the appropriate category-specific discriminator. By doing so, we are able to mitigate the problem of negative transfer, thereby resulting in better overall alignment. MeGA-CDA is evaluated on several benchmark datasets and is shown to outperform existing approaches by a considerable margin. 
\vskip-2.5mm

\section*{Acknowledgement}
This work was supported by the NSF grant 1910141 and Mercedes-Benz Research \& Development India.



{\small
\bibliographystyle{ieee_fullname}
\bibliography{egbib}
}

\end{document}


\title{Supplementary Material for ``MeGA-CDA: Memory Guided Attention for Category-Aware Unsupervised Domain Adaptive Object Detection''}

\maketitle


\section{Network architecture}
In Table~\ref{gcw_fig}, we show the architecture of the global ($\mathcal{D}_{gda}$) and category-wise ($\mathcal{D}^k_{cda}$) discriminators. The architecture of the networks $\Theta_t$ and $\Theta^k_t$ are similar as shown in Table~\ref{feat_ext}.

\begin{table}[!b]
\caption{Ablation analysis of multi-discriminators at instance and image level for Cityscapes $\rightarrow$ Foggy-Cityscapes dataset. k- Number of classes}
\huge
\begin{center}
\resizebox{1.0\linewidth}{!}{
\begin{tabular}{|l|cccccccc|c|}
\hline
Method & person & rider & car & truck & bus & train & mcycle & bicycle & mAP\\
\hline\hline
Source Only & 25.8& 33.7& 35.2& 13.0& 28.2& 9.1& 18.7& 31.4& 24.4\\ \hline
MADA \cite{pei2018multi} & 28.4& 34.7& 35.6& 15.8& 32.1& 10.4& 20.6& 32.3& 26.3\\ \hline
k Multi-disc & 35.5& 44.9& 51.1& 26.5& 40.8& 24.6& 36.3& 37.9& 37.2\\ \hline
GDA+CDA+MA & 36.3& 45.5& 51.7& 25.1& 44.8& 35.8& 32.7& 37.9& 38.7\\ \hline
Oracle & 37.2& 48.2& 52.7& 35.2& 52.2& 48.5& 35.3& 38.8& 43.5\\ \hline
\end{tabular}
}
\end{center}
\label{gda_foggy_cityscape}
\end{table}

\section{Alternative baselines}
Table~\ref{gda_foggy_cityscape} provides additional results corresponding to the ablation analysis of multi-discriminators at the instance and image level. We have employed  MADA \cite{pei2018multi} algorithm in object detection to study the performance of category-wise instance-level domain adaptation. MADA \cite{pei2018multi} is implemented by category-wise domain adaptation loss at instance level multiplied with that predicted class's softmax probability for both source and target domains.   Furthermore, at the image-level, we analyze the global alignment between k-global discriminator and category-wise discriminator. By studying these two methods, we can argue category-wise adaptation is crucial for category-wise alignment for both source and target domain. In the k-global discriminator experiment, k discriminators are applied on the image level. 

From Table~\ref{gda_foggy_cityscape} we can infer that by doing category-wise domain alignment (GDA+CDA+MA) at the image level, mAP is increased by 12\% when compared to category-wise domain alignment at the instance level (MADA). This shows that category-specific alignment approaches used for adapting classification models do not directly translate well for detection models. Furthermore, adding $k$ multi-discriminator at image level perform 2\% lower than category-wise domain alignment at the image level. This shows that blindly adding more discriminators do not have the ability to route specific information through each discriminator to boost overall feature alignment. Adding category-wise $k$ discriminators as proposed in the paper would achieve that by routing only category-specific information to respective discriminators.

\section{Qualitative results}


\begin{table}[t!]
\caption{Global and Category-wise domain discriminator}
\begin{center}
\begin{tabular}{|c|}
\hline
Gradient Reversal Layer \\ \hline
Conv, 1 × 1, 64, stride 1, ReLU \\ \hline
Conv, 3 × 3, 64, stride 1, ReLU \\ \hline
Conv, 3 × 3, 64, stride 1, ReLU \\ \hline
Conv, 3 × 3, 3, stride 1 \\ \hline
\end{tabular}
\end{center}
\label{gcw_fig}
\end{table}

\begin{table}[t!]
\caption{Learned Similarity}
\begin{center}
\begin{tabular}{|c|}
\hline
Conv, 3 × 3, 512, stride 1, ReLU \\ \hline
Conv, 3 × 3, 256, stride 1, ReLU \\ \hline
Conv, 3 × 3, 128, stride 1, ReLU \\ \hline
Conv, 1 × 1, 64, stride 1 \\ \hline
\end{tabular}
\end{center}
\label{feat_ext}
\end{table}

\begin{figure*}
\begin{center}
\includegraphics[width=0.47\linewidth]{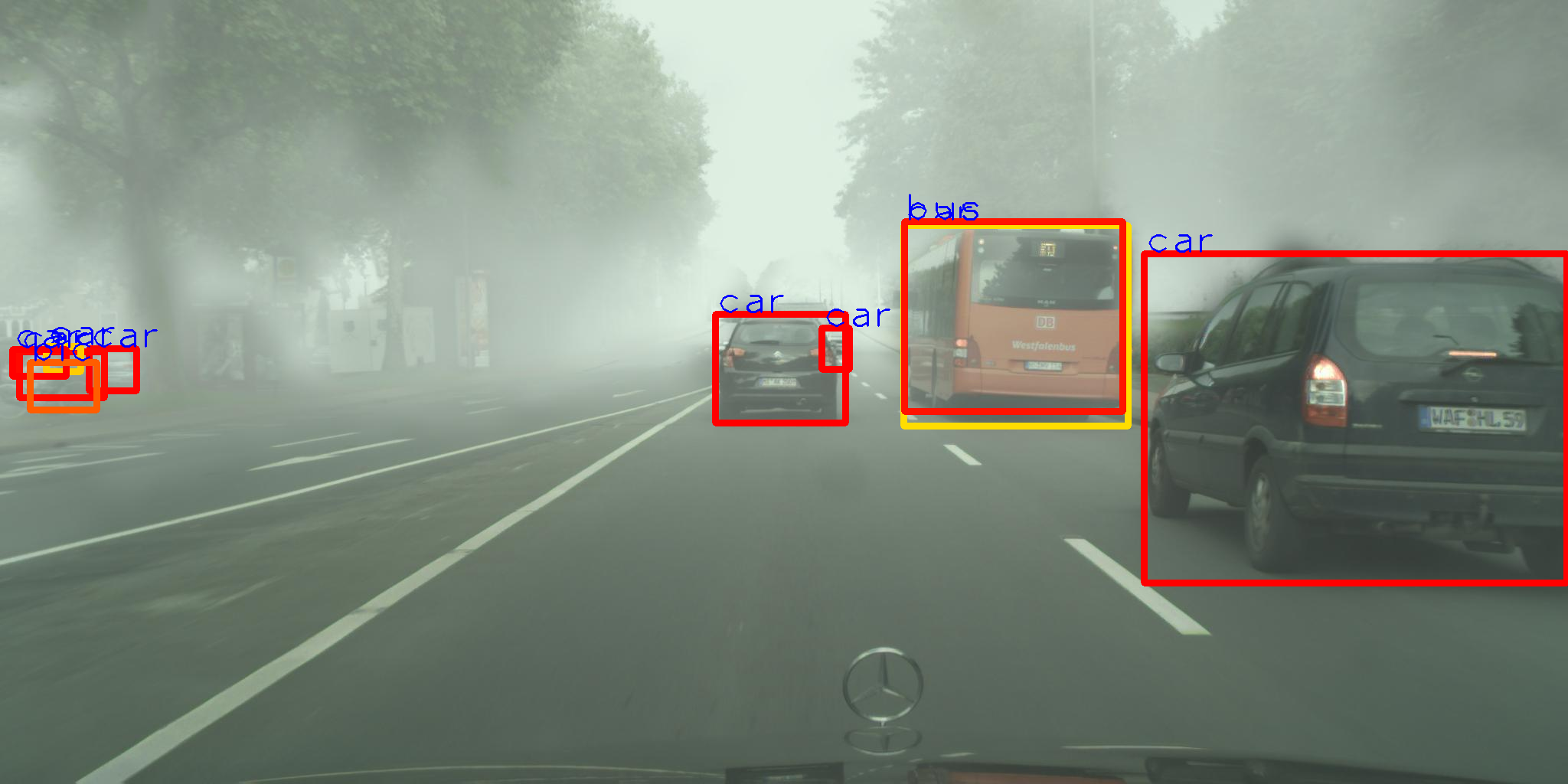}
\includegraphics[width=0.47\linewidth]{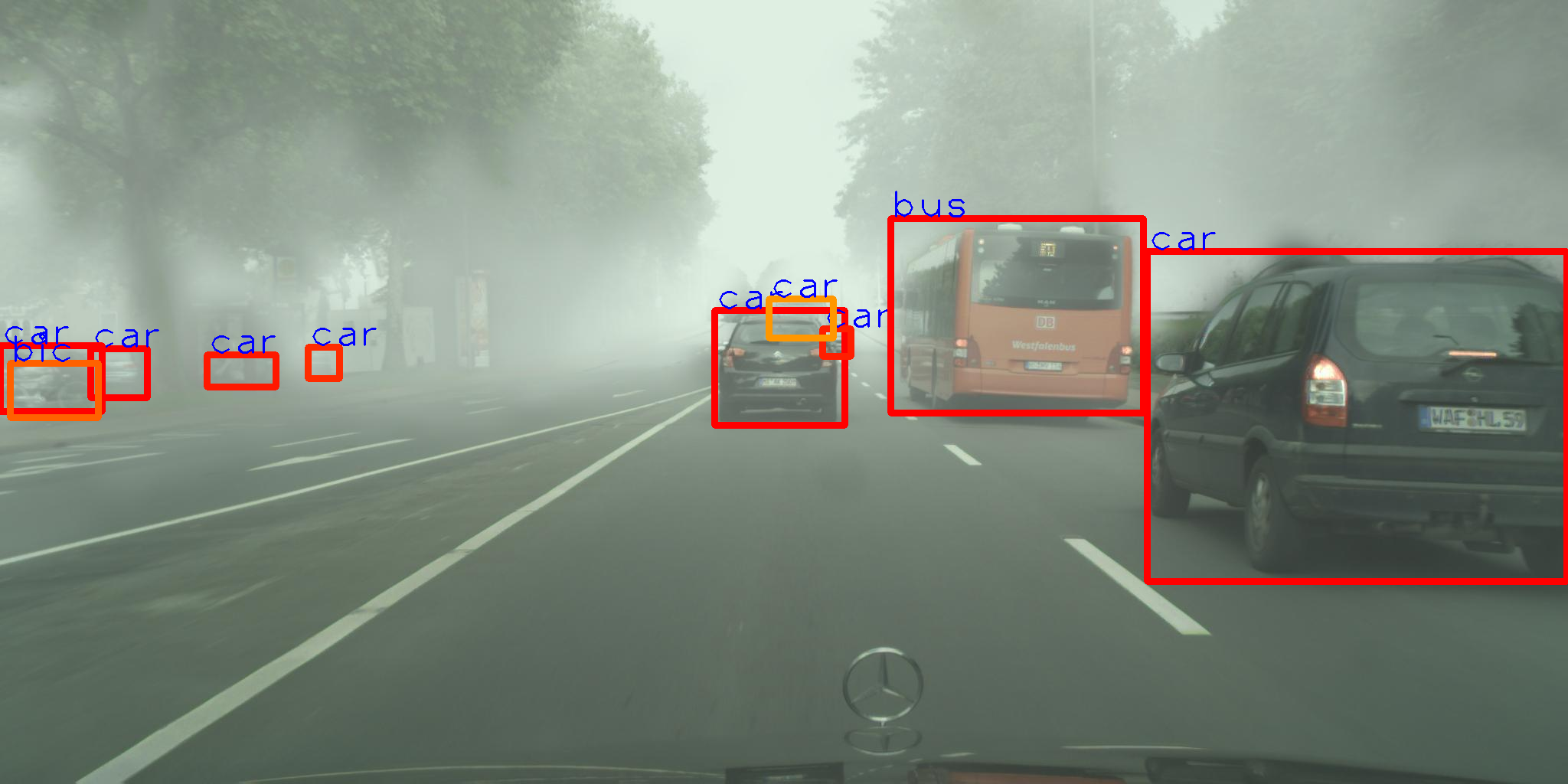}
\includegraphics[width=0.47\linewidth]{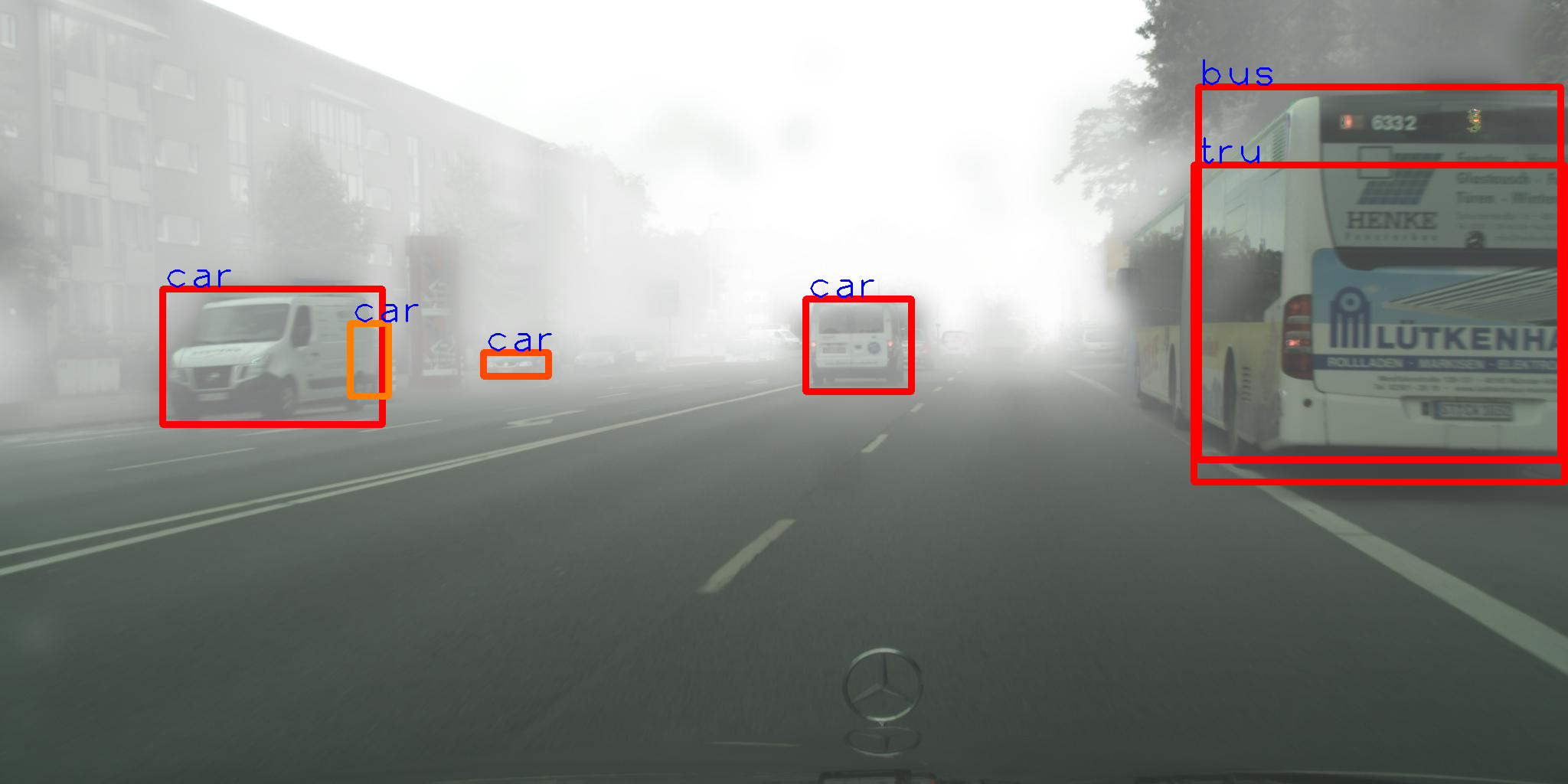}
\includegraphics[width=0.47\linewidth]{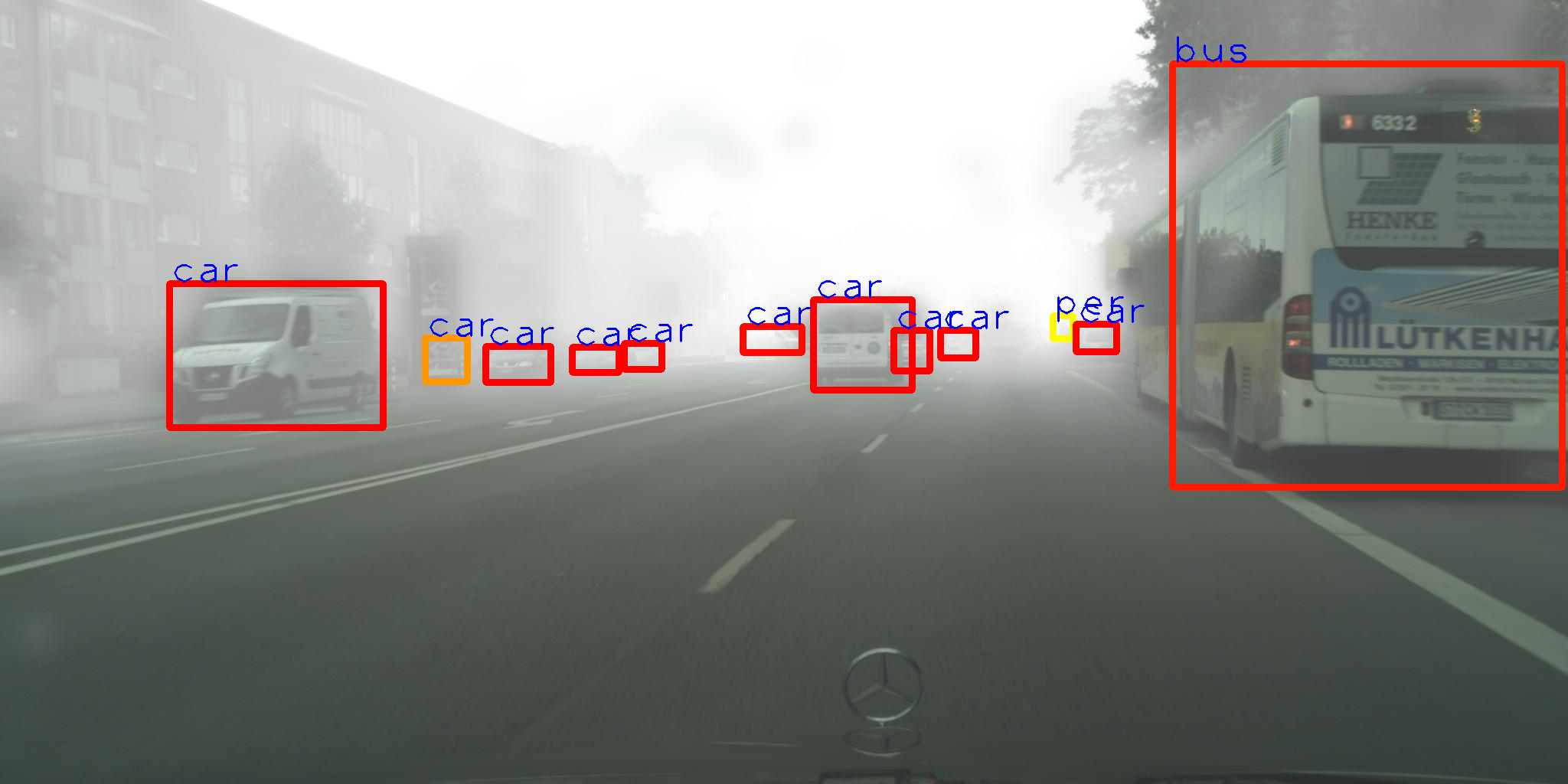}
\includegraphics[width=0.47\linewidth]{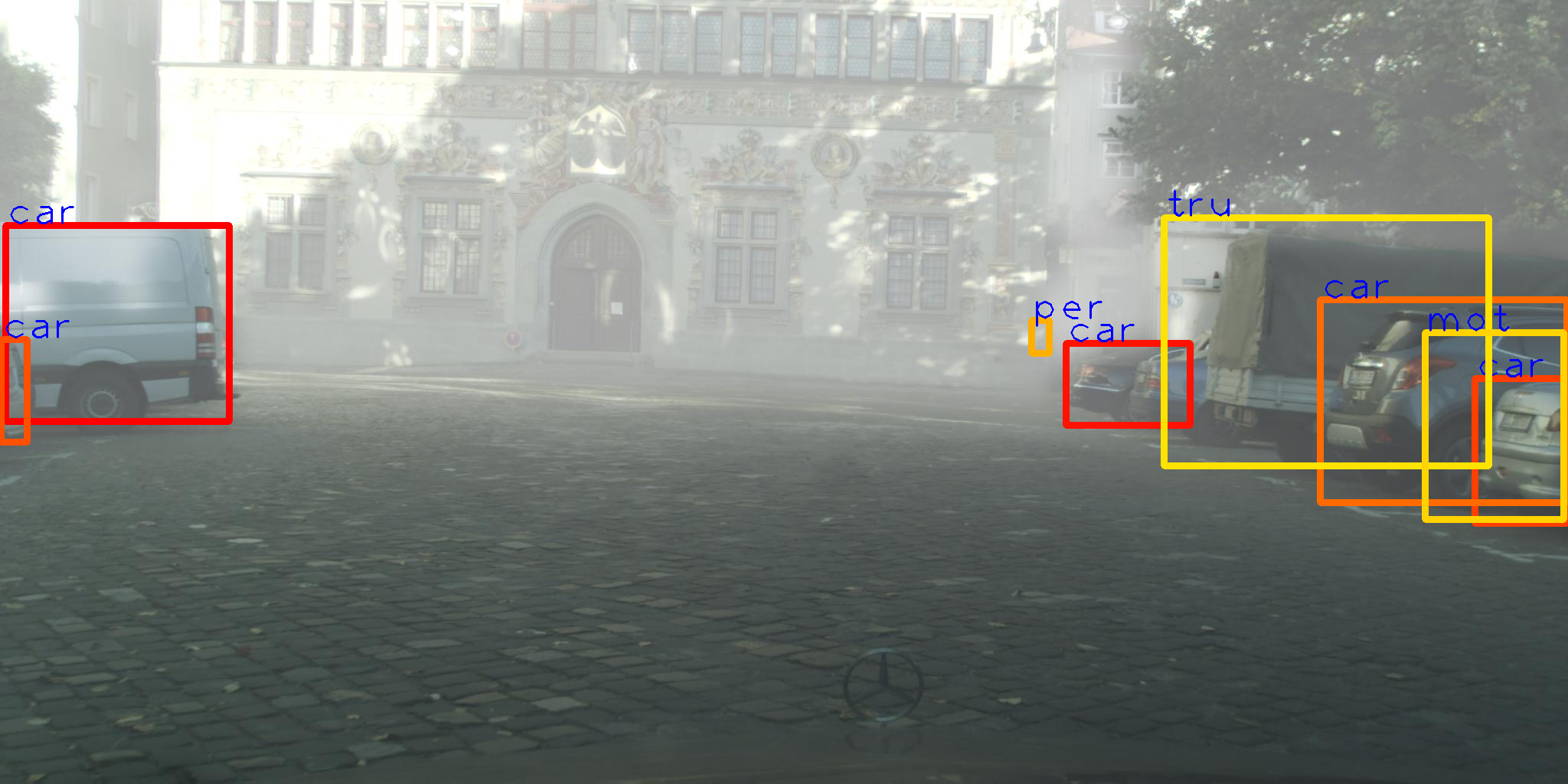}
\includegraphics[width=0.47\linewidth]{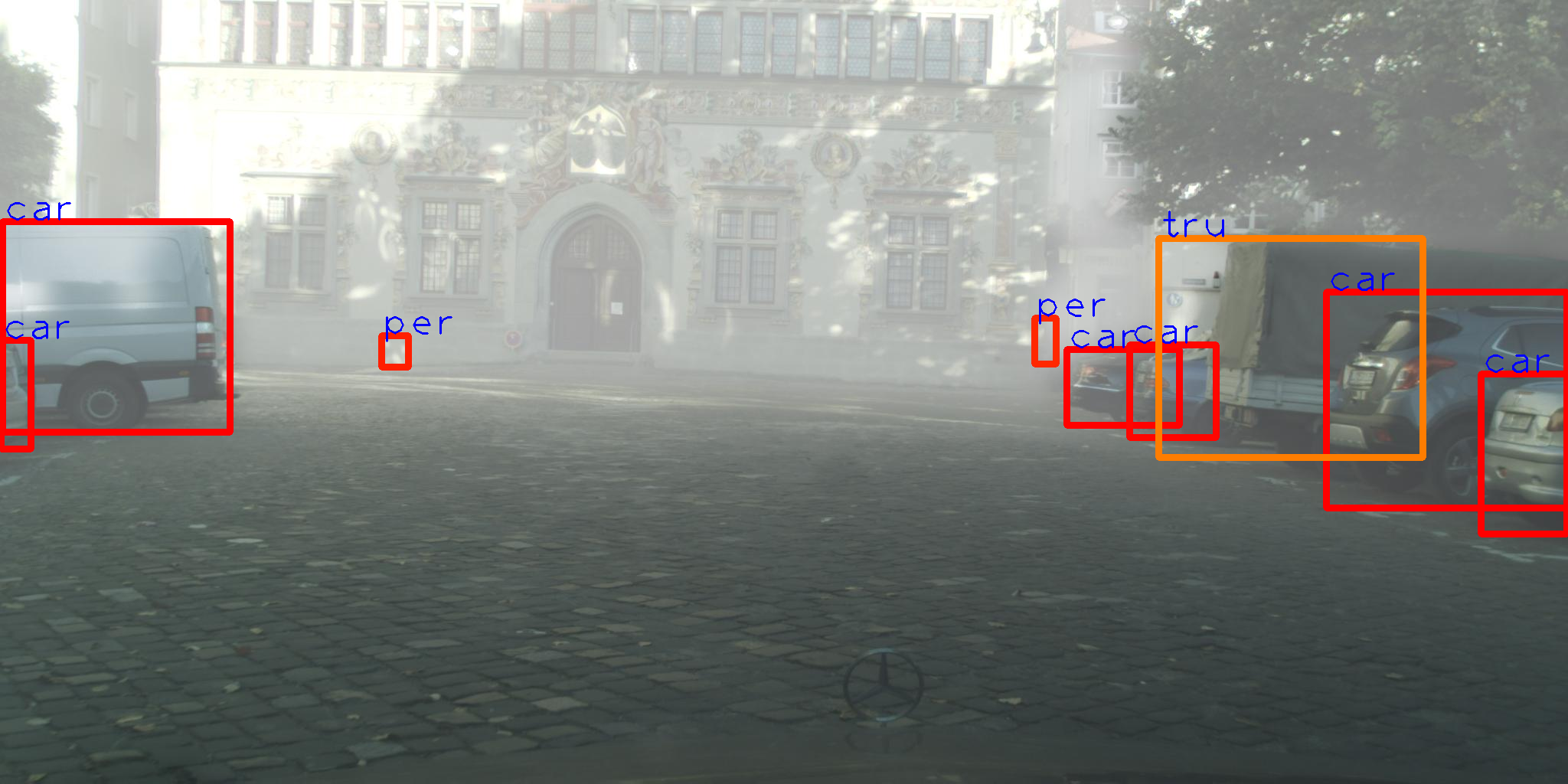}
\includegraphics[width=0.47\linewidth]{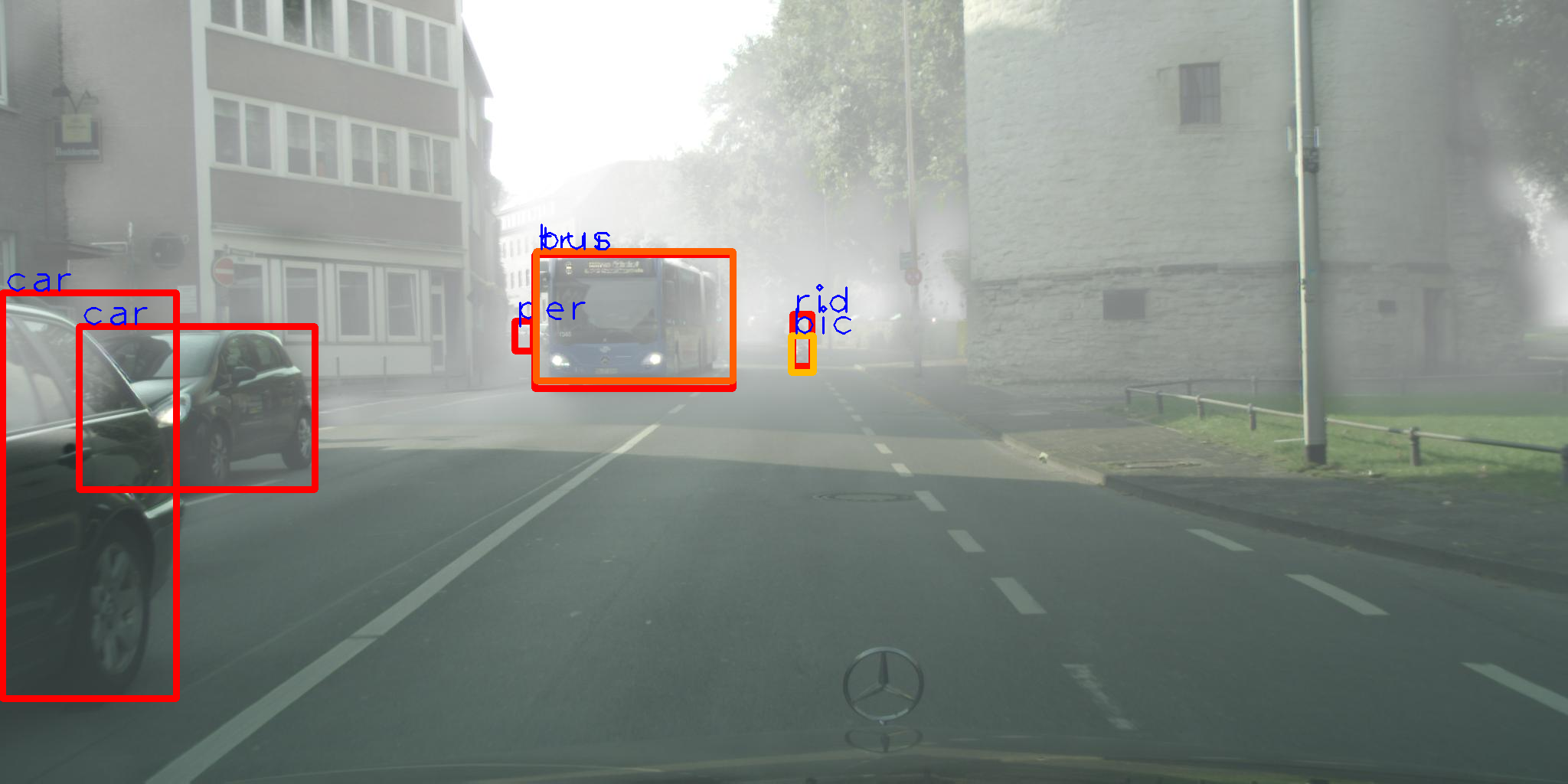}
\includegraphics[width=0.47\linewidth]{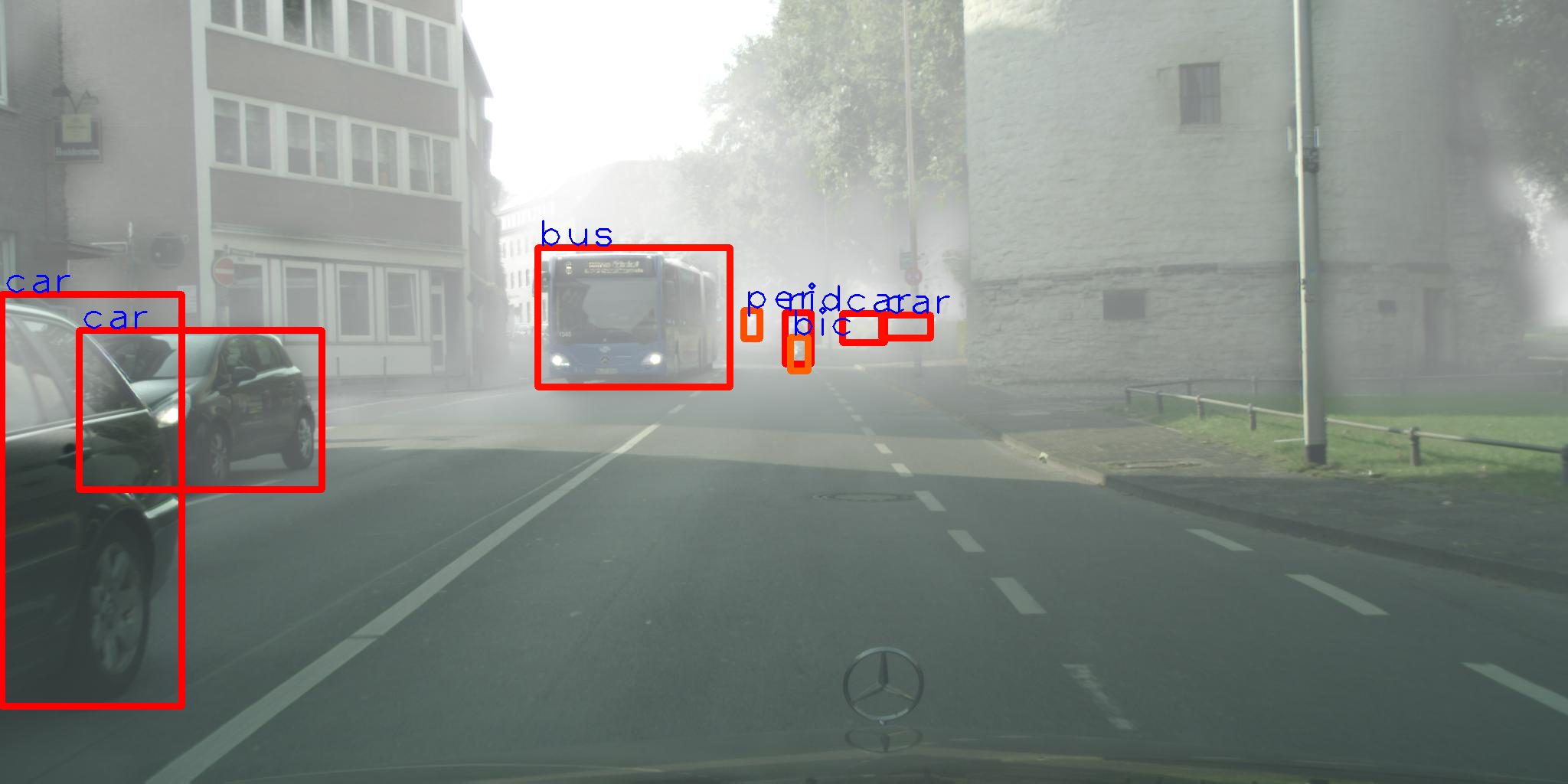}
\includegraphics[width=0.47\linewidth]{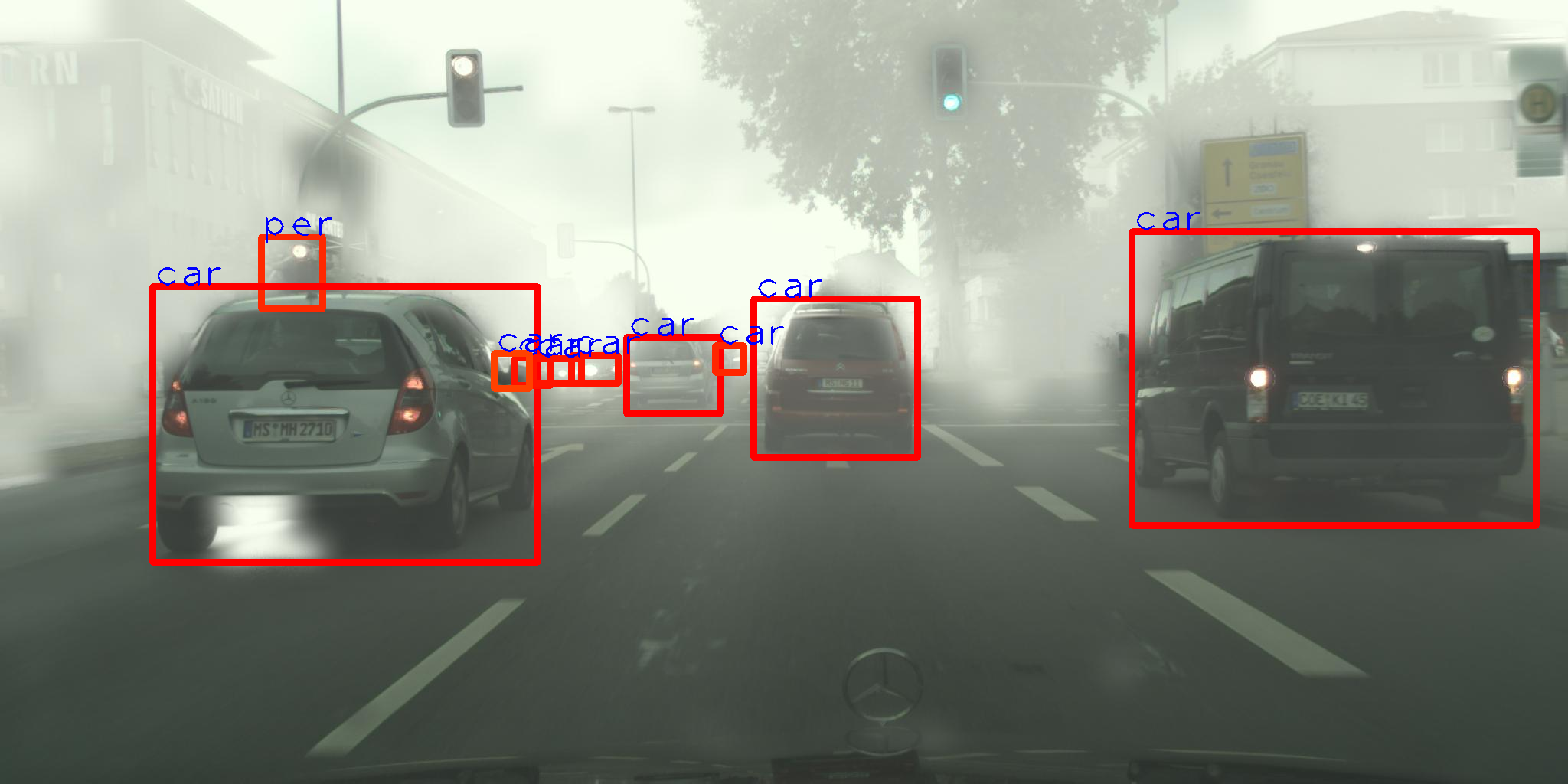}
\includegraphics[width=0.47\linewidth]{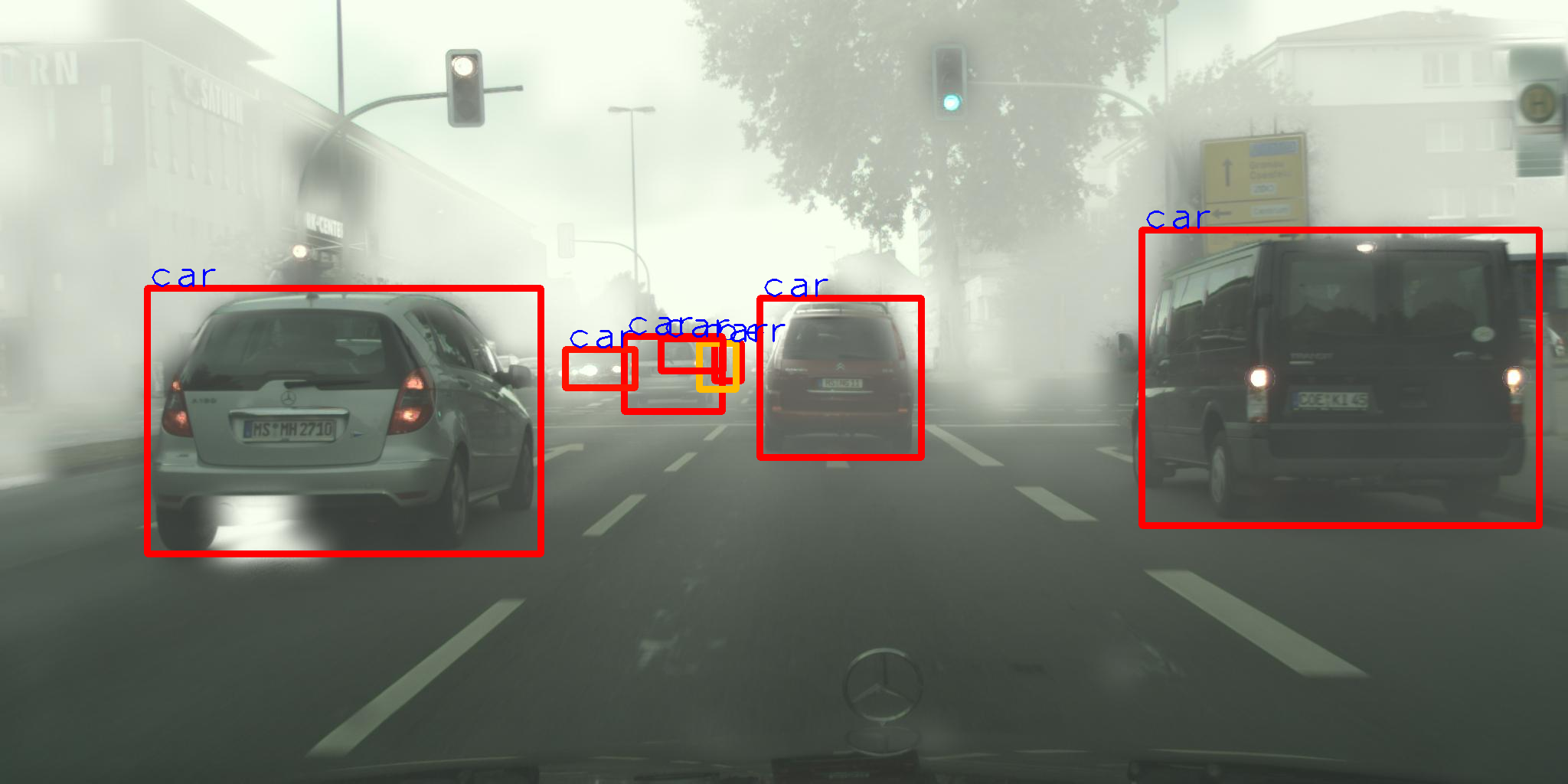}
\includegraphics[width=0.4\linewidth]{Output/score.png}
\vskip -5pt
 DAFaster \hskip 200pt MeGA-CDA 
\end{center}
\vskip -10.0pt \caption{More detection visualization for Foggy$\rightarrow$Cityscapes. Left: DAFaster RCNN, Right: Proposed method. The bounding boxes are colored based on the detector confidence using the color map as shown. From the above visualization, we can infer that our model efficiently tackles classes' negative transfer and constructs high confidence prediction boxes. We show detections with scores higher than 0.5.}
\label{fig:sup_foggy}
\end{figure*}



\begin{figure*}
\begin{center}
\includegraphics[width=0.28\linewidth]{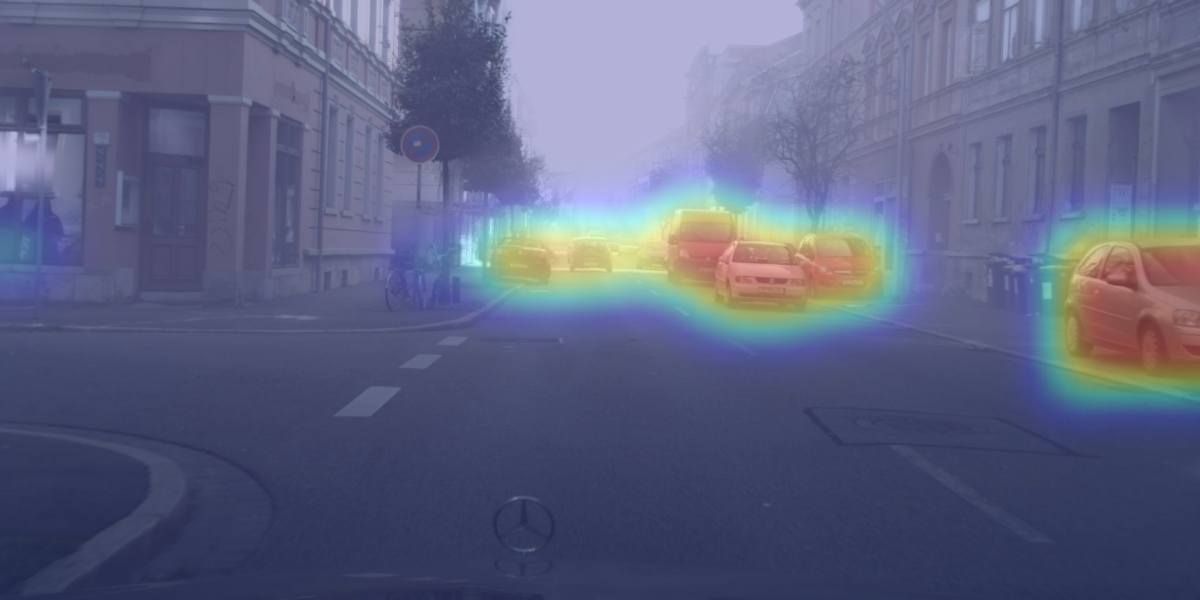}
\includegraphics[width=0.28\linewidth]{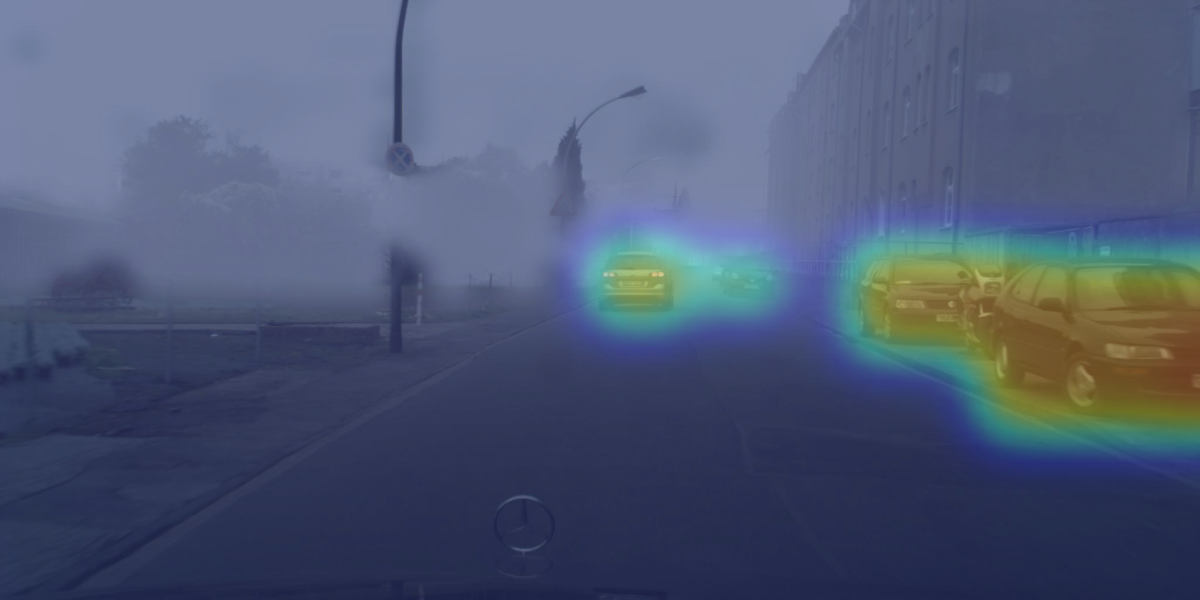}
\includegraphics[width=0.28\linewidth]{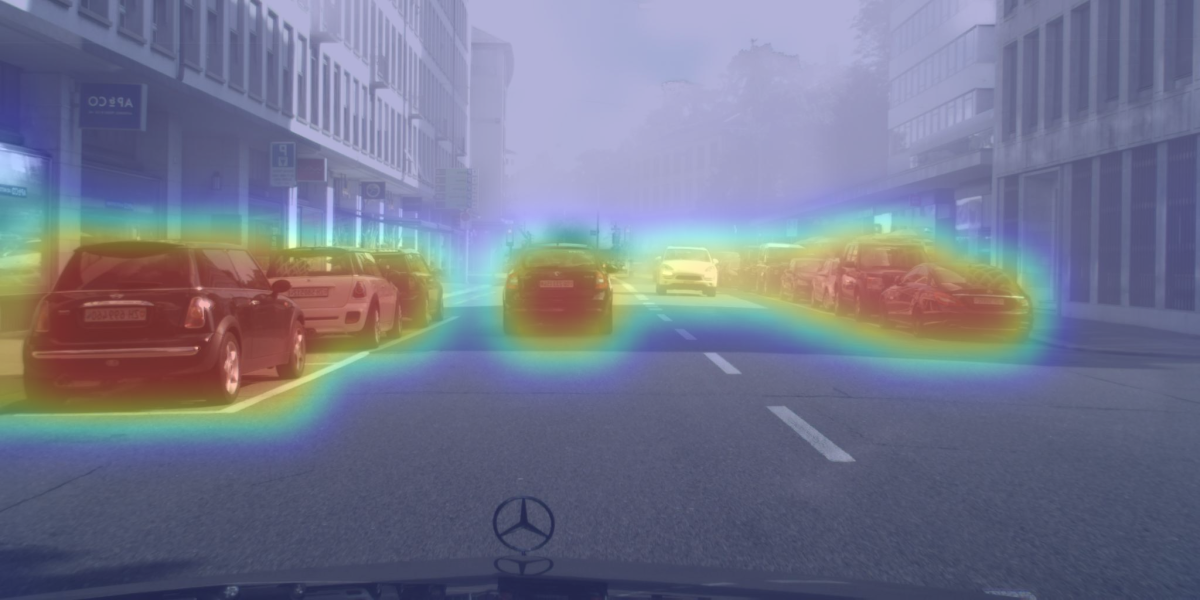}
\includegraphics[width=0.28\linewidth]{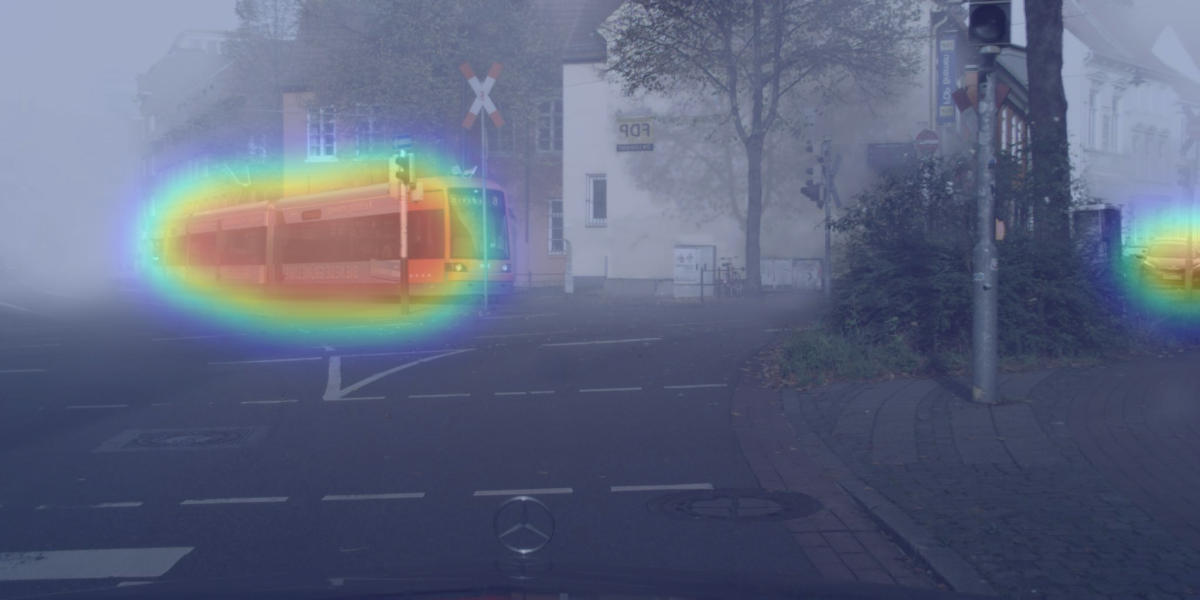}
\includegraphics[width=0.28\linewidth]{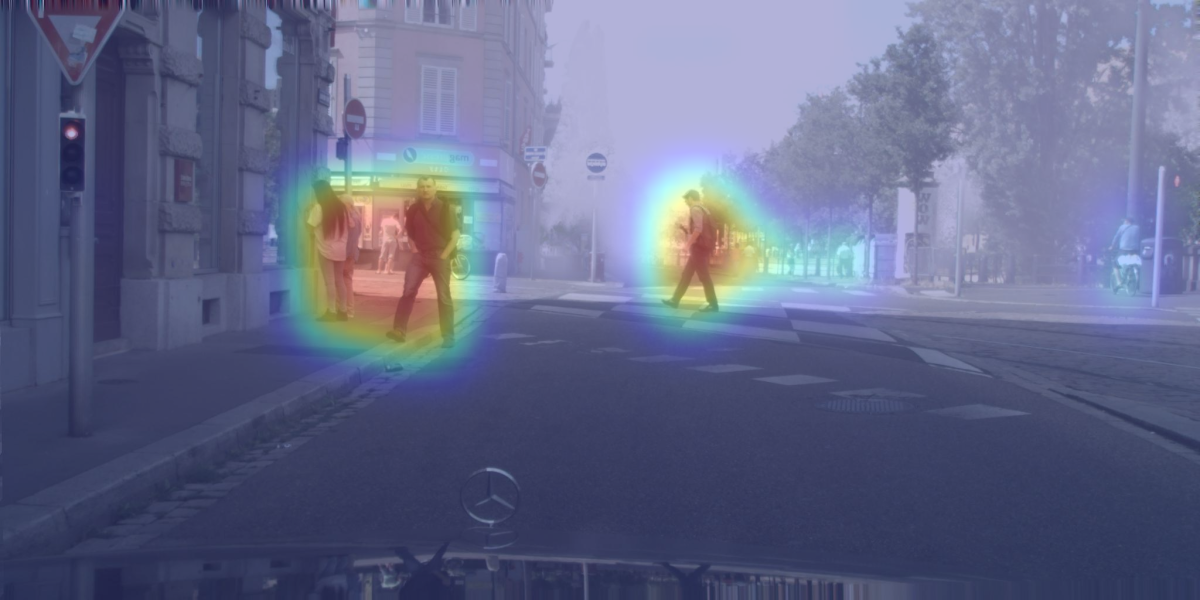}
\includegraphics[width=0.28\linewidth]{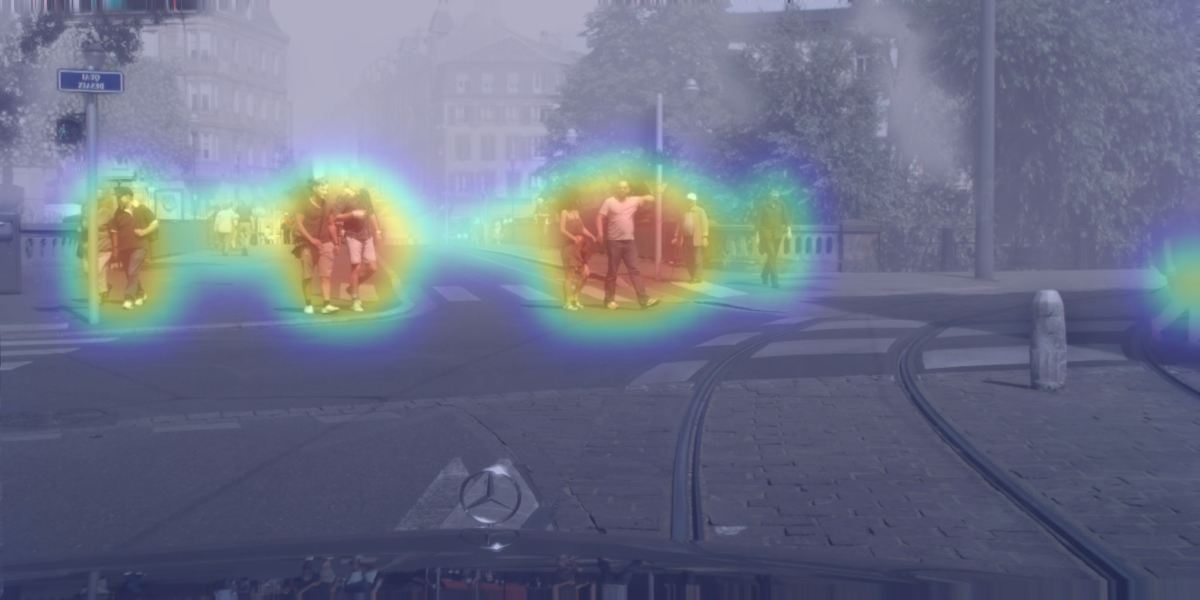}
\includegraphics[width=0.28\linewidth]{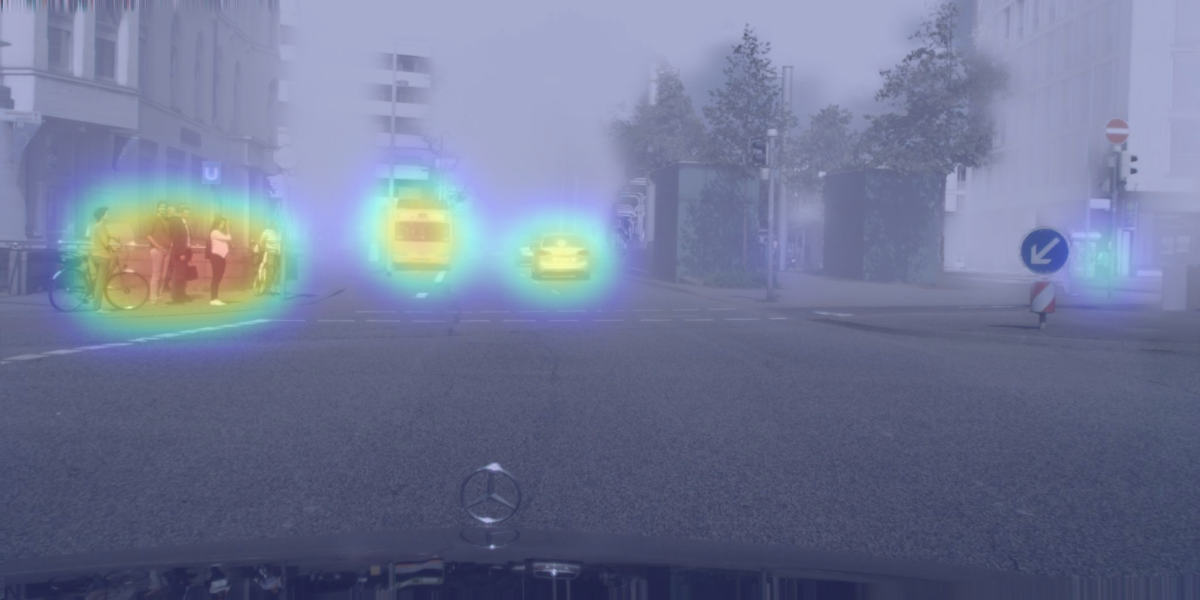}
\includegraphics[width=0.28\linewidth]{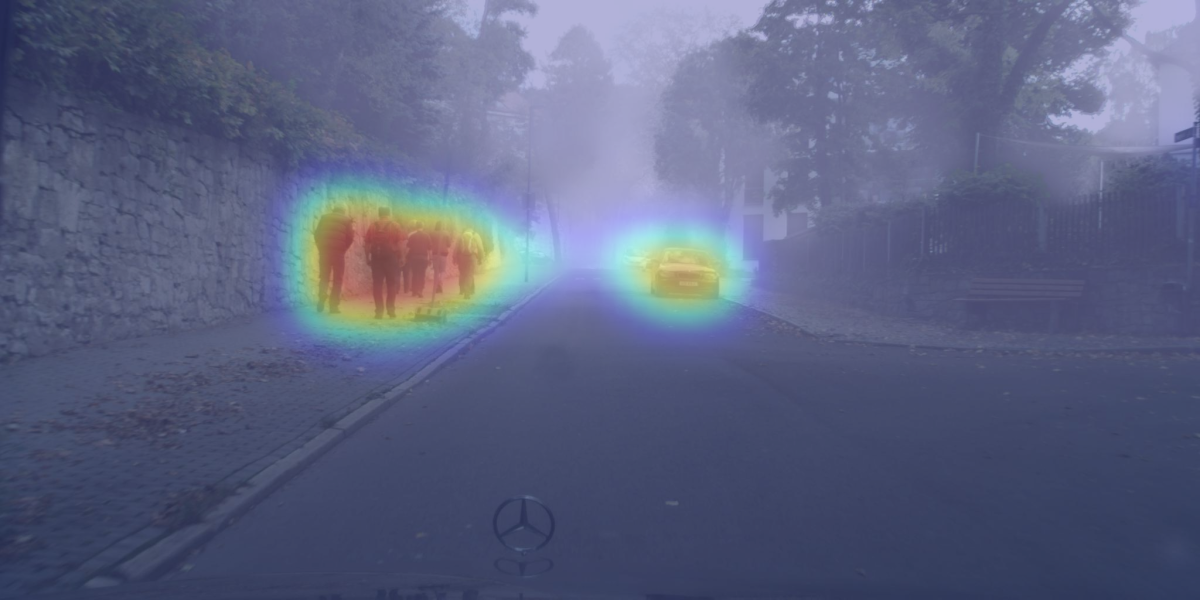}
\includegraphics[width=0.28\linewidth]{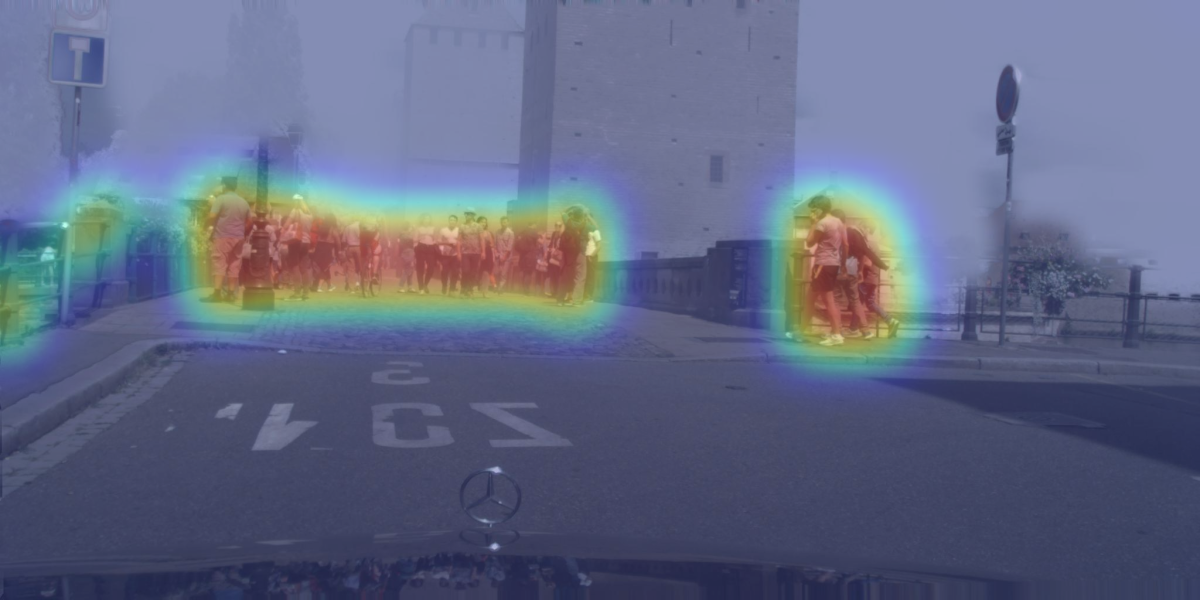}
\includegraphics[width=0.28\linewidth]{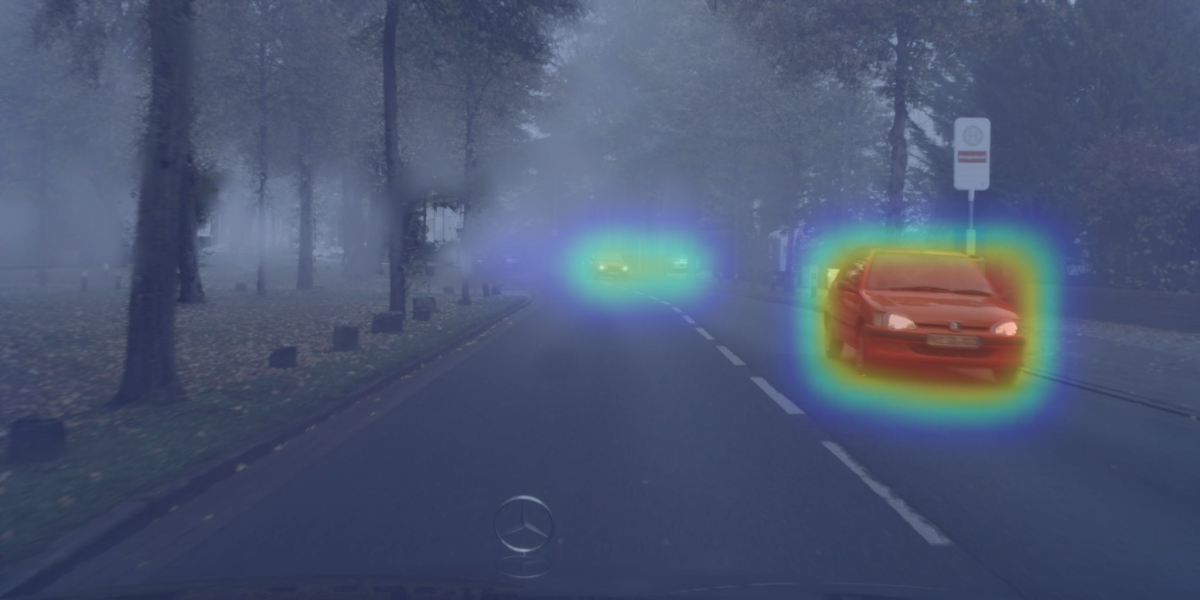}
\includegraphics[width=0.28\linewidth]{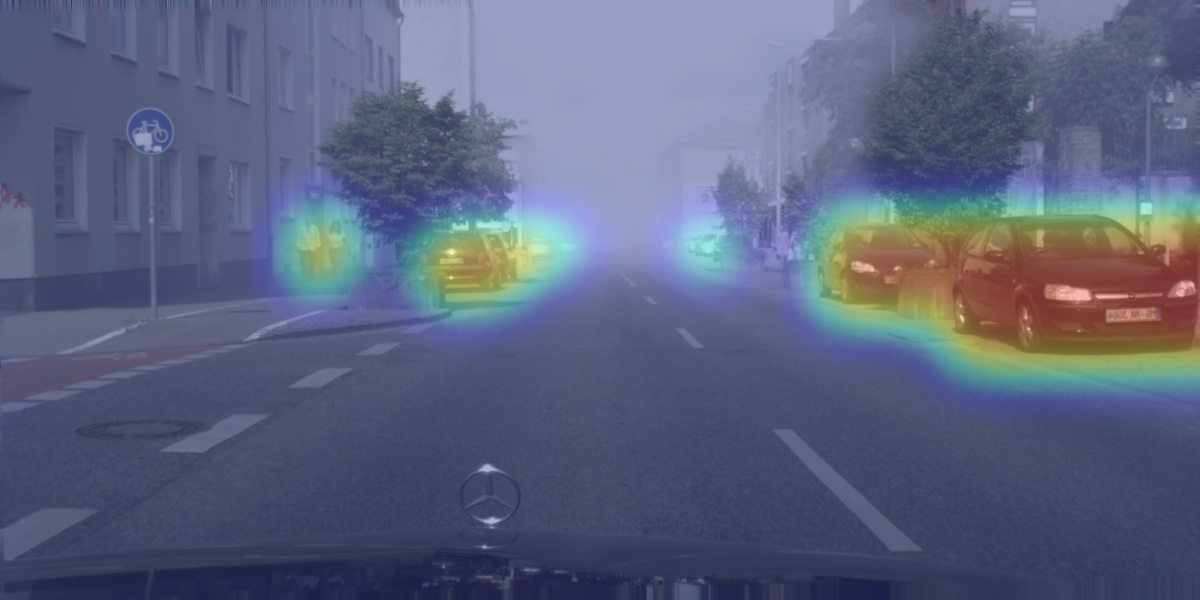}
\includegraphics[width=0.28\linewidth]{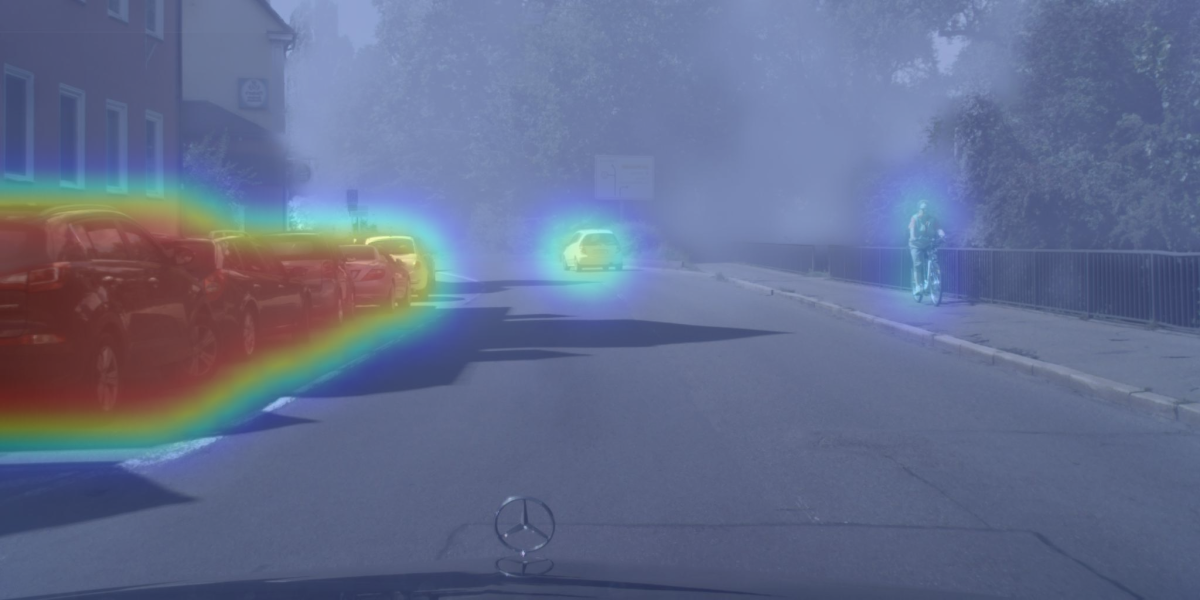}
\includegraphics[width=0.28\linewidth]{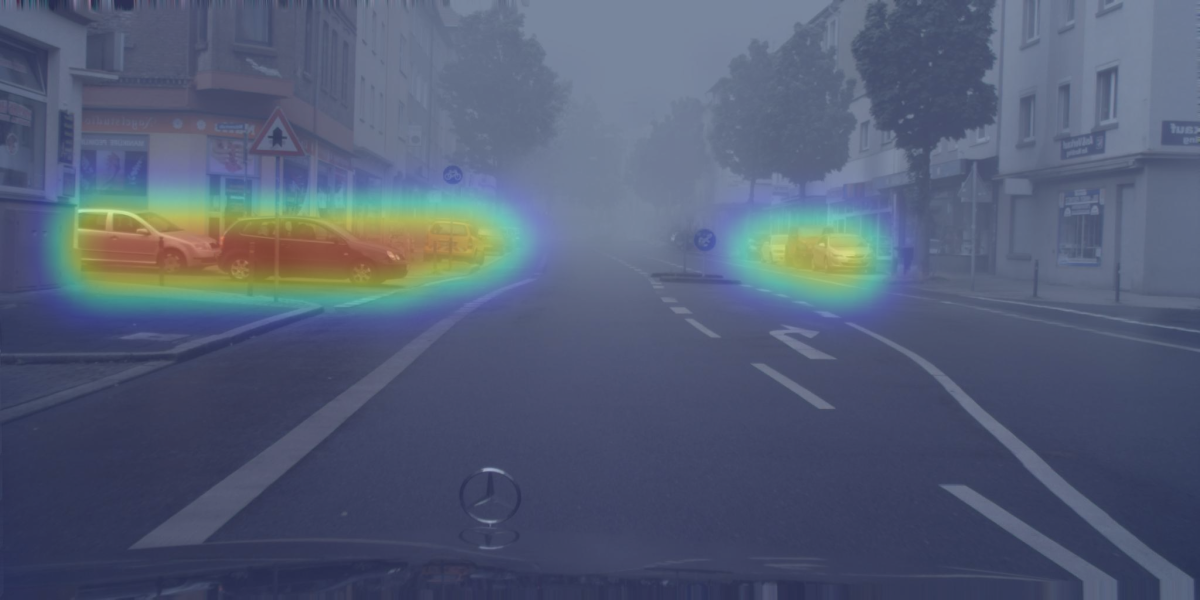}
\includegraphics[width=0.28\linewidth]{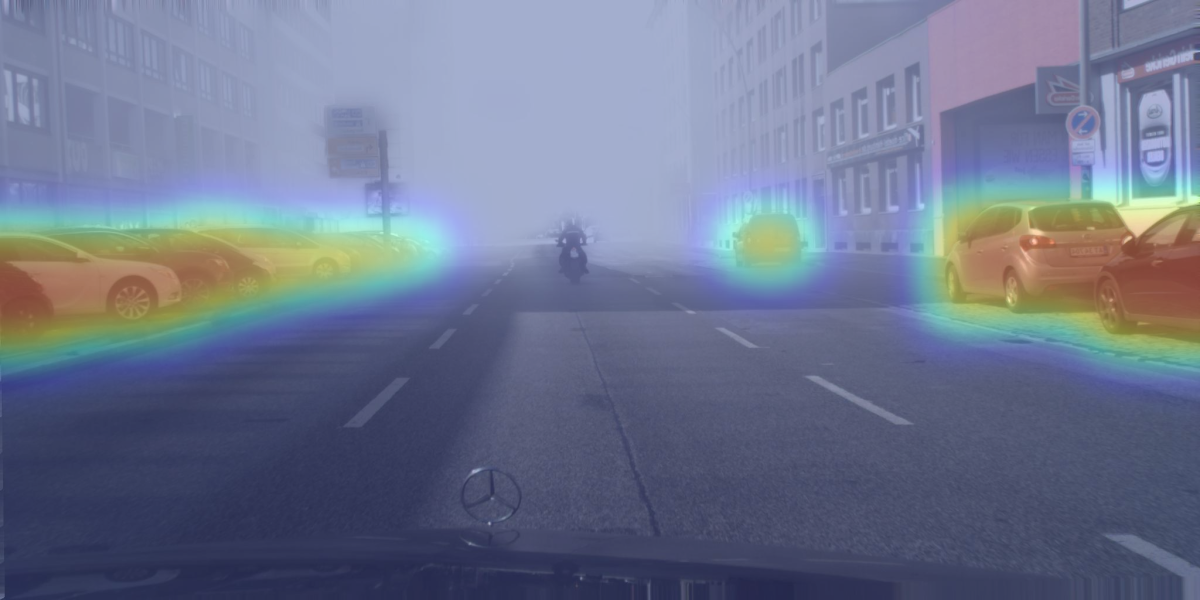}
\includegraphics[width=0.28\linewidth]{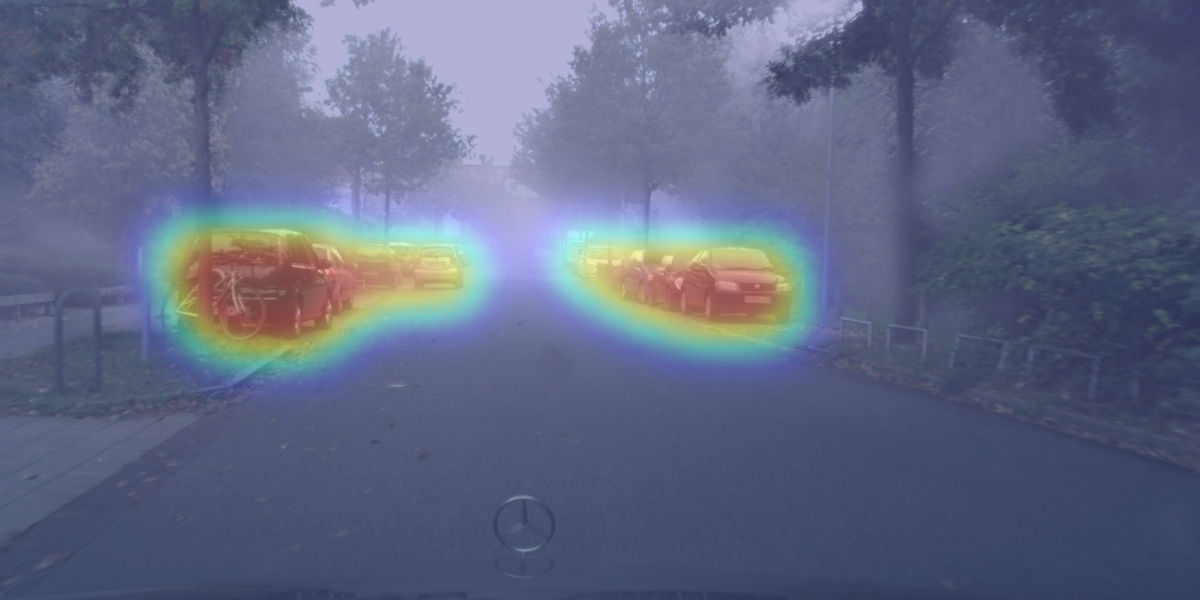}
\end{center}
\vskip -10.0pt \caption{More visualizations of memory guided attention maps on target domains}
\label{fig:sup_mem}
\end{figure*}

{\small
\bibliographystyle{ieee_fullname}
\bibliography{egbib}
}